%% file: main.tex
\documentclass[jair,twoside,11pt,theapa]{article}
\usepackage{jair, theapa, rawfonts}

\jairheading{x}{2020}{1-15}{7/20}{x/20}
\ShortHeadings{A Variational Approach to Unsupervised Sentiment Analysis}
{Zeng, Zhou, Liu, Lin, Song, Kuo, \& Chiu}
\firstpageno{1}

\usepackage{times}
\usepackage{latexsym}

\usepackage{url}
\usepackage{amsmath,amssymb}
\DeclareMathOperator{\E}{\mathbb{E}}

\usepackage{amsfonts}
\usepackage{helvet}
\usepackage{courier}
\usepackage{url}
\usepackage{xcolor}
\usepackage{float,color}
\usepackage{times}
\usepackage{algorithm}
\usepackage{algorithmic}
\usepackage{epsfig}
\usepackage{stfloats}
\usepackage{url}
\usepackage{epstopdf}
\usepackage{multicol}
\usepackage{tocstyle}
\usepackage{dsfont,amsfonts,amssymb,amsmath,color}
\usepackage{multirow}
\usepackage{booktabs}
\usepackage{diagbox}

\def\0{{\bf 0}}
\def\1{{\bf 1}}

\def\JM{{\mathcal J}}
\def\LM{{\mathcal L}}

\def\NM{{\mathcal N}}

\def\PM{{\mathcal P}}

\begin{document}

\title{A Variational Approach to Unsupervised Sentiment Analysis}

\author{\name Ziqian Zeng \email zzengae@cse.ust.hk \\
\addr Department of CSE, Hong Kong University of Science and Technology \\
\AND
\name Wenxuan Zhou \email zhouwenx@usc.edu \\
\addr Department of CS, University of Southern California, CA, USA \\
\AND
\name Xin Liu \email xliucr@cse.ust.hk \\
\name Zizheng Lin \email zlinai@cse.ust.hk \\
\name Yangqiu Song \email yqsong@cse.ust.hk \\
\addr Department of CSE, Hong Kong University of Science and Technology \\
\AND
\name Michael David Kuo \email mikedkuo@gmail.com \\
\name Wan Hang Keith Chiu \email kwhchiu@hku.hk \\
\addr Department of Diagnostic Radiology, University of Hong Kong
}


\maketitle

\input{abstract}
\input{introduction}

\input{method}
\input{experiment}
\input{related_work}

\input{conclusion}

\bibliography{sample}
\bibliographystyle{theapa}

\end{document}

%% file: abstract.tex
\begin{abstract}
In this paper, we propose a variational approach to unsupervised sentiment analysis.
Instead of using ground truth provided by domain experts, we use target-opinion word pairs as a supervision signal. 
For example, in a document snippet ``the room is big,'' (room, big) is a target-opinion word pair.
These word pairs can be extracted by using dependency parsers and simple rules. 
Our objective function is to predict an opinion word given a target word while our ultimate goal is to learn a sentiment classifier. 
By introducing a latent variable, i.e., the sentiment polarity, to the objective function, we can inject the sentiment classifier to the objective function via the evidence lower bound. 
We can learn a sentiment classifier by optimizing the lower bound. 
We also impose sophisticated constraints on opinion words as regularization which encourages that if two documents have similar (dissimilar) opinion words, the sentiment classifiers should produce similar (different) probability distribution. 
We apply our method to sentiment analysis on customer reviews and clinical narratives. 
The experiment results show our method can outperform unsupervised baselines in sentiment analysis task on both domains, and our method obtains comparable results to the supervised method with hundreds of labels per aspect in customer reviews domain, and obtains comparable results to supervised methods in clinical narratives domain. 
\end{abstract} 

%% file: introduction.tex
\section{Introduction}
Sentiment analysis is a task of identifying the sentiment polarity expressed in textual material. 
Sentiment Analysis is an important and useful application in natural language processing \cite{liu2012sentiment}. 
For example, sentiment analysis on customer reviews can help business owners to improve their products and help other customers to purchase products wisely. 
Sentiment analysis on clinical narratives which contains personal observations and attitudes from nurses, radiologists, or therapists, can provide automated decision support for physicians. 
Most existing sentiment analysis methods are supervised methods which requires labeled training data. 
In practice, naturally labeled data is scarce and annotating data could be expensive. 
Although crowd-sourcing is a cost-saving way to annotate data, some tasks still require domain experts such as doctors, lawyers, or financial consultants. 
Hence, sentiment analysis in the unsupervised setting is a more realistic scenario than supervised one.

We aim to tackle the challenge of unsupervised sentiment analysis. 
A traditional way to perform unsupervised sentiment analysis is lexicon-based methods \cite{turney2002thumbs,maite2011lexicon}. 
A lexicon consists of a list of opinion words and corresponding sentiment orientation (SO) scores capturing polarity (positive or negative) and strength (degree to which the opinion word is positive or negative). 
Then, for any given text, all opinion words are extracted and annotated with their SO scores using a lexicon. 
The SO scores are aggregated into a single score for the text. 
There are two disadvantages of lexicon-based methods. 
First, unsupervised lexicon-based methods do not involve any training process. 
Second, the performance of these methods is highly dependent on the quality of pre-defined lexicons. 
Most of lexicons suffer from either high coverage/low precision or low coverage/high precision problem \cite{ng2006examining}.

We propose an unsupervised method to train a sentiment classifier without relying on the quality of the lexicon. 
The supervision signal of our method is pairs of words.
We observe that normally there is a syntactic dependency between a opinion word and a target word. 
A target word is usually related to a subjective. 
An opinion word is related to a sentiment polarity. 
For example, given a document snippet ``the hotel is comfortable,'' in sentiment classification task which aims to classify the sentiment polarity of a hotel, ``comfortable'' is an opinion word and ``hotel'' is a target word. 
The objective function of our algorithm is to maximize the likelihood of an opinion word given a target word. 
By introducing a latent variable, i.e., the sentiment polarity of a document, to the objective function, we can inject a sentiment classifier to the objective function via the evidence lower bound (ELBO). 
There are two classifiers in the ELBO, i.e, a sentiment classifier and an opinion word classifier. 
The input of the sentiment classifier is a document representation, and the inputs of opinion word classifier are a possible value of sentiment polarity and a target word. 
The sentiment classifier produces a probability distribution, which is used to approximate the true posterior distribution in ELBO. 
Then, we can learn a sentiment classifier (an approximated posterior distribution) by optimizing the lower bound. 
Further to make use of opinion words, we also apply a regularization term which encourages that if two documents have similar (dissimilar) opinion words, the sentiment classifiers should produce similar (different) probability distribution. 
We apply our algorithm on two domains, i.e., sentiment analysis on customer reviews and radiology reports. 
In customer reviews domain, our task is document-level multi-aspect sentiment classification (DMSC). 
The goal of DMSC is to predict the sentiment polarity (e.g., positive or negative) of each aspect given a document in which there are several sentences and each sentence describes one or more aspects. 
The objective function is to predict an opinion word such as ``good'' given a target word such as ``price.'' 
In clinical domain, our task is hip fracture classification. 
The goal of hip fracture classification is to predict the heath status (e.g., fracture or non-fracture) given a textual radiology report which describes the health status of the hip region. 
The objective function predict an opinion word such ``fractured'' given a target word such as ``trochanter''.

Our framework has two advantages. First, in realistic scenario, we could manually define target words to perform sentiment analysis with different granularity.  
For example, if we want to do a coarse-grained sentiment classification on hotel reviews, target words would be \{price, view, room, location, ...\}. While if a fine-grained classification is needed, e.g., the Internet quality, then target words would be \{Internet, WiFi, access, connection, ...\}. 
The supervision signal is very flexible in realistic scenario where the sentiment classification tasks are in different granularity. 
Second, under our framework, the sentiment classifier would be any neural network architectures which have demonstrated high capacity in supervised learning setting. The upper bound of the performance of our sentiment classifier is that of the same model in supervised settings. 
Our framework will benefit from powerful neural networks.

Our contribution can be summarized as follows:

$\bullet$ We propose an unsupervised approach to solve sentiment analysis.

$\bullet$ We propose to learn a sentiment classifier by injecting it into another relevant objective via the evidence lower bound. This framework is flexible to adopt different neural network architectures as sentiment classifiers and perform classification tasks with different granularity.

$\bullet$ The experiment results show our method can outperform unsupervised baselines in both tasks and obtains comparable results to the supervised method with hundreds of labels per aspect in DMSC task, and comparable results to supervised methods in hip fracture classification task.

A preliminary version of this work \cite{zeng2019variational} appeared in the proceedings of NAACL 2019. This journal version has made several major improvements. 
First, our method is generalized to sentiment classification task. Previous work \cite{zeng2019variational} focuses on document-level multi-aspect sentiment classification task.  
Second, we add more details of the methodology section. We discuss different forms of training objective when applying different assumptions. We also introduce approximation techniques to deal with the situations that the opinion vocabulary is large or the number of categories is large. 
Third, we add a regularization term to encourage that if two documents have similar (dissimilar) opinion words, the sentiment classifiers should produce similar (different) probability distribution. 
Last, we add more experiments on clinical sentiment analysis and conduct a case study.

The remainder of the paper is organized as follows. 
We first introduce our method including sentiment classifier, opinion word classifier, training objective and regularization term in Section 2. 
Then we shows our experiments including target opinion word pars extraction, compared methods,  result and error analysis in Section 3. 
Finally, we introduce the related work in Section 4 followed by the conclusion of this paper in Section 5.

%% file: method.tex
\section{Methodology}
In this section, we introduce our variational approach to unsupervised sentiment analysis. 
We will introduce sentiment classifier, opinion word classifier, training objective, and regularization of opinion word respectively in the following subsections.


\begin{figure}
    \center
    \includegraphics[width=1.0\textwidth]{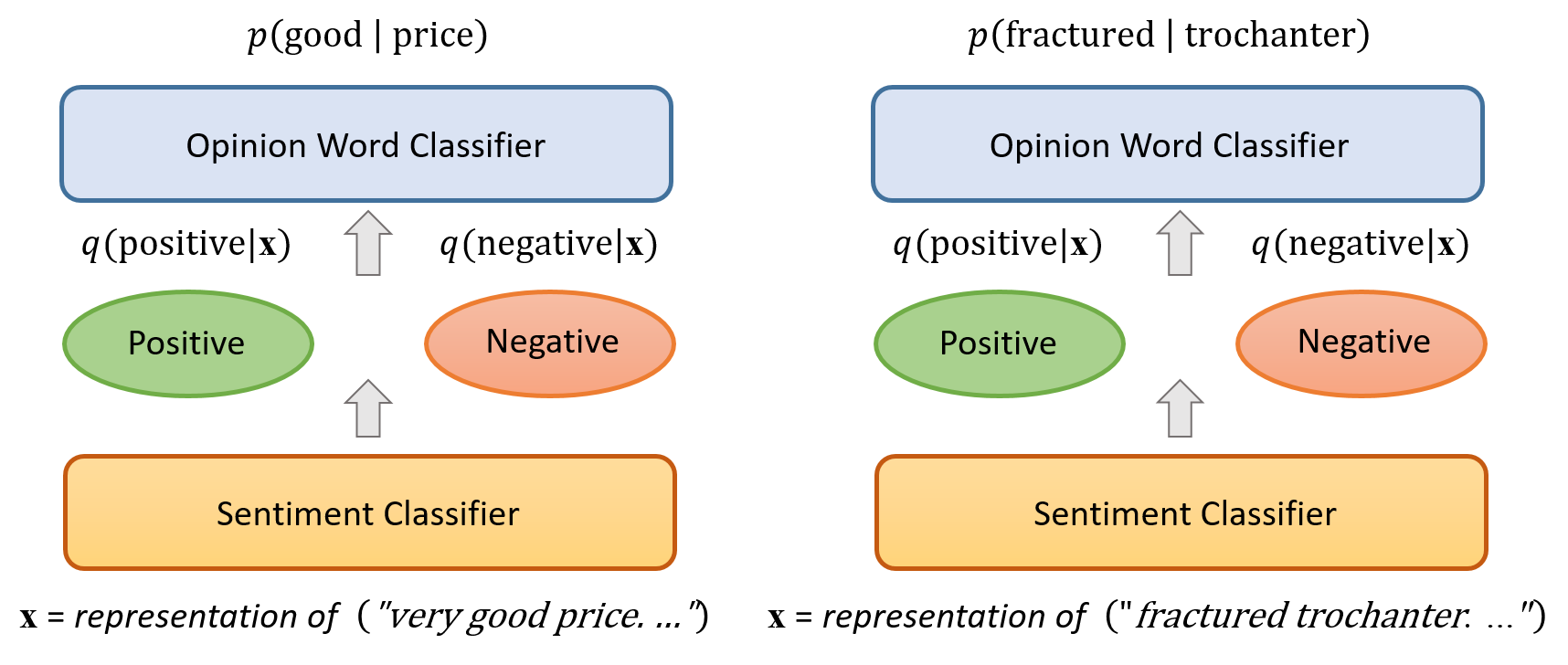}\vspace{-0.05in}
    \caption{Our model consists of a sentiment classifier and an opinion word classifier.}\label{fig:model-illustrate}
\end{figure}

\subsection{Overview} 
Our model consists of a sentiment classifier and an opinion word classifier. 
Our goal is to learn a sentiment classifier to predict the sentiment polarity of a document. 
It could be the sentiment polarity of an aspect given a user review or the health status of the hip given textual radiology report. 
The input of the sentiment classifier is the text representation of a user review or a radiology report. 

Figure \ref{fig:model-illustrate} shows the relation between two classifiers. 
The input $\mathbf{x}$ of the sentiment classifier is the text representation of a user review or a radiology report, e.g., bag-of-words or a representation learned by neural networks.
Let $C$ be a random variable, indicating the sentiment polarity of a document, e.g., the sentiment polarity of an aspect or the health status of the hip. 
The sentiment classifier takes $\mathbf{x}$ as input and produces a probability distribution, denoted as $q(C|\mathbf{x})$. 
For example, if $C$ has two possible values, i.e., positive and negative, then outputs of the classifier are $q(positive|\mathbf{x})$ and $q(negative|\mathbf{x})$ respectively. 
The opinion word classifier of DMSC task takes a target word (``price'') and a possible value of sentiment polarity $c$ as input, and estimates $p(\text{``good''}|c,\text{``price''})$.
The opinion word classifier of hip fracture classification task takes a target word (``trochanter'') and a possible value of health status $c$ as input, and estimates $p(\text{``fractured''}|c,\text{``trochanter''})$. 
 Our training objective function is to maximize the log-likelihood of an opinion word given a target word, e.g., $p(\text{``good''}|\text{``price''})$ or $p(\text{``fractured''}|\text{``trochanter''})$. 
The ELBO of log-likelihood consists of a sentiment classifier and an opinion word classifier.


\subsection{Sentiment Classifier}
The sentiment classifier aims to estimate a distribution $q(C|\mathbf{x})$, where $C$ is a discrete random variable representing the sentiment polarity of a document, e.g., the sentiment polarity or the health status of the hip, and $\mathbf{x}$ is a feature representation of a document. 
Let $c$ denote a possible value of the random variable $C$, representing a possible value of sentiment polarity, e.g., \textit{positive}, \textit{negative}, \textit{fractured} or \textit{non-fractured}. 
The sentiment classifier estimates the probability as
\begin{equation}\label{eq:encoder-softmax}
q(C = c|\mathbf{x}) = \frac{\exp \big( \mathbf{w}_{c}^{T} \mathbf{x} \big)}{\sum_{c'}{\exp \big( \mathbf{w}_{c'}^{T} \mathbf{x} \big)}} \; ,
\end{equation}
where $\mathbf{w}_{c}$ is a vector associated with a sentiment polarity $c$.

The representation of a document $\mathbf{x}$ can be various. 
Traditional document representations of text classification would be bag-of-words, n-gram, or averaged word embeddings. 
Recently, end-to-end neural network based models demonstrate a powerful capacity to extract features of a document \cite{devlin2018bert}. Our model will benefit growingly powerful neural network based feature extraction methods.
We use bag-of-words, convolutional neural networks, or long short-term memory networks as the document representation methods in our experiments.


\subsection{Opinion Word Classifier}
The opinion word classifier aims to estimate the probability of an opinion word $w_o$ given a target word $w_t$ and a possible value of sentiment polarity $c$:
\begin{equation}
p(w_o |c, w_t) = \frac{\exp \big( \varphi ({w_o}, {w_t}, c) \big)}{ \sum_{{w_o'}}{ \exp \big( \varphi ({w_o'}, {w_t}, c) \big) } } \;,
\label{dwc}
\end{equation}
where $\varphi(\cdot)$ is a scoring function which takes opinion word $w_o$, target word $w_t$, and a possible value of sentiment polarity $c$ as inputs. 
The nature of the score function is about the frequency of occurrence. 
If an opinion word, a target word, and a possible value of sentiment polarity co-occur frequently, the score will be high, otherwise, it will be low. 
The scoring function could be various, and here we use the simplest one (dot product) as the scoring function:
\begin{equation}
\varphi ({w_o}, {w_t}, c)= I ( w_t \in \mathcal{K} ) \cdot \mathbf{a}_{c}^T \mathbf{w}_o \;,
\end{equation}
where $\mathbf{w}_o$ is the word embedding of opinion word $w_o$, $\mathbf{a}_{c}$ is a vector associated with $c$,  $\mathcal{K}$ is the set of relevant target words, and
$I(\cdot)$ is an indicator function where $I(true)=1$ and $I(false)=0$. 
The scoring function could be various, e.g., multilayer perceptron (MLP). Here we only introduce the simplest case. 

Given a target word $w_t$ and a possible value of sentiment polarity $c$, we aim to maximize the probability of opinion words highly related to them.
For example, the opinion word ``good'' is usually related to the target word ``price'' for the aspect \textit{value} with sentiment polarity \textit{positive}, and the opinion word ``terrible'' is usually related to the target word ``traffic'' for the aspect \textit{location} with sentiment polarity \textit{negative}. 
The opinion word ``fractured'' is usually related to the target word ``trochanter'' with the health status of hip \textit{positive}, and the opinion word ``intact'' is usually related to target word ``trochanter'' with the health status of hip \textit{negative}. 

\subsection{Training Objective}
The objective function is to maximize the log-likelihood of an opinion word $w_o$ given a target word $w_t$. 
After introducing a latent variable (i.e., the sentiment polarity of a document) to the objective function, we can derive an evidence lower bound (ELBO) of the log-likelihood which can incorporate two classifiers. 
The first one corresponds to the sentiment classifier.
The second one corresponds to the opinion word classifier. 
The ELBO of log-likelihood is shown as follows,
\begin{align}
\LM & = \sum_{x \in X} \sum_{(w_o,w_t) \in \PM_{x}} \log p(w_o | w_t) \nonumber \\
& = \sum_{x \in X} \sum_{(w_o,w_t) \in \PM_{x}} \log \sum_{c}p(w_o , c | w_t) \nonumber\\
& = \sum_{x \in X} \sum_{(w_o,w_t) \in \PM_{x}} \log \sum_{c} q(c|\mathbf{x}) \Big[ \frac{p(w_o , c | w_t)}{q(c|\mathbf{x})} \Big] \nonumber \\
& \geq \sum_{x \in X} \sum_{(w_o,w_t) \in \PM_{x}} \sum_{c} q(c|\mathbf{x}) \Big[ \log \frac{p(w_o , c | w_t)}{q(c|\mathbf{x})} \Big] \nonumber \\
& = \sum_{x \in X} \sum_{(w_o,w_t) \in \PM_{x}} \E_{q(C|\mathbf{x})} \big[\log p(w_o|c,w_t)p(c|w_t) \big]  + H(q(C|\mathbf{x})), \label{vari}
\end{align}
where $X$ is the training set containing all documents, and $\PM_{x}$ is the set of all word pairs extracted from a document $x$, $H(\cdot)$ refers to the Shannon entropy, and $q(c|\mathbf{x})$ is short for $q(C=c|\mathbf{x})$. By applying Jensen's inequality, the log-likelihood is lower-bounded by Eq. (\ref{vari}). 

The equality holds if and only if the KL-divergence of two distributions, $q(C|\mathbf{x})$ and $p(C|w_t,w_o)$, equals to zero. 
Maximizing the evidence lower bound is equivalent to minimizing the KL-divergence. 
The proof is shown as follows. 
To be concise, the proof only consider one training sample.
\begin{align}
\log p(w_o | w_t) & = \sum_{c} q(c|\mathbf{x}) \log  p(w_o | w_t) \nonumber\\
& = \sum_{c} q(c|\mathbf{x}) \log \frac{ p(w_o |w_t) q(c|\mathbf{x}) p(w_o,c | w_t) }{  p(w_o,c|w_t) q(c|\mathbf{x}) } \nonumber\\
& = \sum_{c} q(c|\mathbf{x}) \log \frac{ p(w_o,c|w_t) }{ q(c|\mathbf{x}) } + \sum_{c} q(c|\mathbf{x}) \log \frac{ p(w_o |w_t) q(c|\mathbf{x})}{  p(w_o,c | w_t) }\nonumber\\
& = LB + \sum_{c} q(c|\mathbf{x}) \log \frac{q(c|\mathbf{x})}{ p(c|w_t,w_o) } \nonumber\\
& = LB + KL\big(q(C|\mathbf{x}) || p(C|w_t,w_o)\big),
\end{align}
where $LB$ is short for Eq. (\ref{vari}) with one training sample. 
Hence, we can learn a sentiment classifier that can produce a similar distribution to the true posterior $p(C|w_t,w_o)$. 
Compared with $p(C|w_t,w_o)$, $q(C|\mathbf{x})$ is more flexible since it can take any kind of feature representations of whole document as input. It is possible for $q(C|\mathbf{x})$ to approximate the true posterior distribution, because a document includes the target word and the opinion word.

We assume that a target word $w_t$ and a possible value of sentiment polarity $c$ are independent, i.e., $p(c|w_t)$ = $p(c)$, since the polarity assignment is not influenced by the target word. For example, only given a target word ``price'', the probability of \textit{positive} and \textit{negative} are equal. By this assumption, the ELBO becomes, 
\begin{align}
    \LM_1 & = \sum_{x \in X} \sum_{(w_o,w_t) \in \PM_{x}} \E_{q(C|\mathbf{x})} \big[\log p(w_o|c,w_t)p(c) \big]  + H(q(C|\mathbf{x})), \label{ob2}
\end{align}
For piror distribution of sentiment polarity, i.e., $p(C)$, we could use a neural network to parameterize it. 
Or we could assume $C$ follows a uniform distribution, which means $p(c)$ is a constant.  
If we assume $C$ follows a uniform distribution, we would remove $p(c)$ in Eq. (\ref{vari}) to get a new objective function as follows, 
\begin{align} 
\LM_2 & = \sum_{x \in X} \sum_{(w_o,w_t) \in \PM_{x}} \E_{q(C|\mathbf{x})} \left[\log p(w_o|c,w_t) \right] + H(q(C|\mathbf{x})) \;. \label{ob3}
\end{align}

\subsection{Approximation}
The partition function in Eq. (\ref{dwc}) requires the summation over all opinion words in the vocabulary. 
If the size of the opinion word vocabulary is large, we could use the negative sampling technique \cite{mikolov2013distributed} to approximate Eq. (\ref{dwc}).
Specifically, we approximate $p(w_o|c,w_t)$ in the objective (\ref{dwc}) with the following objective function:
\begin{equation}
\log \sigma \big( \varphi ({w_o}, {w_t}, c) \big) + \sum_{ w_o' \in \NM } \log\big( 1 - \sigma \big(\varphi ({w_o'}, {w_t}, c) \big) \big) \; \label{app1},
\end{equation}
where $w_o'$ is a negative sample in opinion words vocabulary, $\NM$ is the set of negative samples and $\sigma(\cdot)$ is the sigmoid function. 
If the approximation is used, we should multiply a hyper-parameter $\alpha$ to the entropy term to ensure that the approximation part and entropy term into are on the same scale~\cite{marcheggiani2016discrete}.
The objective function becomes,
\begin{align} 
\LM_3 & = \sum_{x \in X} \sum_{(w_o,w_t) \in \PM_{x}}  \E_{q(C|\mathbf{x})} \big[ \log \sigma \big( \varphi ({w_o}, {w_t}, c) \big) + \sum_{ w_o' \in \NM } \log \big( 1 - \sigma \big( \varphi ({w_o'}, {w_t}, c) \big) \big) + \log p(c|w_t) \big] \nonumber \\
\quad \quad &+\sum_{x \in X} \sum_{(w_o,w_t) \in \PM_{x}} \alpha H(q(C|\mathbf{x}))\; \label{app1}.
\end{align}
The expectation operation in Eq. (\ref{vari}) requires the summation over all possible sentiment polarities. 
If the number of polarities is large, we could use likelihood ratio gradient estimators \cite{glynn1987likelilood} to approximate the expectation part. 
Using the likelihood ratio estimation, the gradient of $\LM$ with respect to parameters $\theta$ in $q(C|\mathbf{x})$ can be computed as follows. To be concise, we only consider one training sample.
\begin{align}
    \nabla_{\theta} \LM = \E_{q(C|\mathbf{x};\theta)} [A(w_o,w_t,c) \nabla \log q(c|\mathbf{x})  ], \label{app2}
\end{align}
where $A(w_o,w_t,c)  = \log p(w_o|c,w_t) + \log p(c|w_t) - \log q(c|\mathbf{x})$ can be treated as the learning signal in policy gradient. 
The gradient in Eq. (\ref{app2}) can be approximated by the Monte Carlo method using $K$ samples of the latent variable: 
\begin{align}
    \nabla_{\theta} \LM \approx \frac{1}{K} \sum_{j=1}^{K} A(w_o,w_t,j) \nabla \log q(C=j|\mathbf{x}) . \label{app3}
\end{align}

\subsection{Regularization on Opinion Words}
The motivation of regularization is that if the opinion words extracted from two documents are similar semantically, then these two documents probably are in the same cluster, and if the opinion words are opposite semantically, then these two documents are probably not in the same cluster. 
For example, in the sentiment classification task, if one document contains a word pair (staff, friendly) and the other one contains a word pair (waiter, unfriendly), it is highly possible that these two documents do not belong to the same cluster because ``friendly'' and ``unfriendly'' are opposite semantically. 
In hip fracture classification, if one document contains a word pair (trochanter, displaced) and the other one contains a word pair (hip, fractured), it is highly possible that these two documents belong to the same cluster because ``displaced'' and ``fractured'' are similar semantically.

First, we will measure the semantic similarity between two opinion words by introducing a score function. 
The score function takes the cosine similarity between the embeddings of two words $w^{i}_{d}$ and $w^{j}_{d}$ as inputs, 
\begin{align}
  s(w^{i}_{d}, w^{j}_{d}) =  \tau \big( \cos(\mathbf{w}^{i}_{d},\mathbf{w}^{j}_{d}) ; \gamma \big),
\end{align}
where $w^{i}_{d}$ and $w^{j}_{d}$ are the opinion word in document $x^i$ and $x^j$ respectively, $\mathbf{w}^{i}_{d}$ and $\mathbf{w}^{j}_{d}$ are corresponding embeddings, ${\gamma} \in [0,1]$ is a threshold, and $\tau(z;\gamma)=z$ if $|z| \geq {\gamma}$ otherwise $\tau(z;\gamma)=0$. 
There are two cases that the score function is zero.
The first one is that the cosine value is positive and is less than a threshold $\gamma$. 
The second one is that the cosine value is negative and is greater than a threshold $-\gamma$. 
The first case means the two opinion words are not similar enough. 
The second case means that two opinion words are not dissimilar enough. 
In both cases, the regularization will not be applied. 

Then we maximize the following objective function: 
\begin{align}
  R(x^i,x^j) &= -d\big(q(C|\mathbf{x}^i), q(C|\mathbf{x}^j)\big)s(w^{i}_{d}, w^{j}_{d}),
\end{align}
where $d\left( \cdot, \cdot \right)$ is squared Euclidean distance, i.e., $d\big( q(C|\mathbf{x}^i), q(C|\mathbf{x}^j) \big) = \sum_{c} \big(q(c|\mathbf{x}^i) - q(c|\mathbf{x}^j)\big)^2$ , and $\mathbf{x}^i$ and $\mathbf{x}^j$ are documents representation of $x^i$ and $x^j$ respectively. 
If the posterior distributions $q(C|\mathbf{x}^i)$ and $q(C|\mathbf{x}^j)$ are different from each other, i.e., $d\big( q(C|\mathbf{x}^i), q(C|\mathbf{x}^j) \big)$ is large, but opinion words suggest that these two documents should be in the same cluster i.e., $s(w^{i}_{d}, w^{j}_{d})$ is large, then $d\big( q(C|\mathbf{x}^i), q(C|\mathbf{x}^j) \big)$ will be encouraged to be small by applying the regularization. 
If applying the regularization, the objective function becomes, 

\begin{align} \label{final-ob}
\JM & = \LM + \beta \sum_{x^i \in X} \sum_{x^j \in X} R(x^i,x^j)\;,
\end{align}
where $x^i$ and $x^j$ are documents in the training set $X$.
The constraints are defined in a $|X| \times |X|$ space. 
In practice, we train our model batch by batch. 
So we only apply the constraints within a mini-batch. 
There are at most $|X_b| \times |X_b|$ constraints in a mini-batch, where $|X_b|$ is the number of samples in a mini-batch. 

%% file: experiment.tex
\section{Experiment}
In this section, we report average sentiment classification accuracies over all aspects on binary document level multi-aspect sentiment classification task and the accuracy on binary hip fracture classification task.
\subsection{Datasets}
\subsubsection{Document Level Multi-aspect Sentiment Analysis} 
We evaluate our model on TripAdvisor \cite{wang2010latent} and BeerAdvocate \cite{mcauley2012learning,lei2016rationalizing,yin2017document} datasets, which contain seven aspects (value, room, location, cleanliness, check-in or front desk, service, and business) and four aspects (feel, look, smell, and taste) respectively.
We run the same preprocessing steps as \cite{yin2017document}.
Both datasets are split into train/development/test sets with proportions 8:1:1. 
All methods can use development set to tune their hyper-parameters.

Ratings of TripAdvisor and BeerAdvocate datasets are on scales of $1$ to $5$ and $0$ to $5$ respectively.
But in BeerAdvocate, $0$ star is rare, so we treat the scale as $1$ to $5$.
We convert original scales to binary scales as follows: $1$ and $2$ stars are treated as negative, $3$ is ignored, and $4$ and $5$ stars are treated as positive.
In BeerAdvocate, most reviews have positive polarities, so to avoid the unbalanced issue, we perform data selection according to overall polarities.
After data selection, the number of reviews with negative overall polarities and that with positive overall polarities are equal.

\subsubsection{Hip Fracture Classification} 
The dataset is collected from Hong Kong Hospital Authority which is a territory-wise and multi-center database storing abundant medical data such as clinical summary, radiology report, and radiology image.
Hong Kong Hospital Authority has collected medical data since 1999. This dataset is collected during the period of 2008 – 2017. 
In this dataset, each sample consists of multiple radiology images and a textual radiology report. 
We only use textual radiology reports in our experiment. 
All samples come from two patient groups, one of which comprises $|X_p|$ (number of positive samples) patients with hip fracture(s), and the other group consists of $|X_n|$ (number of negative samples) patients aged $65$ or above with hip or pelvic X-ray conducted and with no hip fracture diagnosed. 
The whole dataset is highly unbalanced.
The number of negative samples (no hip fracture) is much larger than that of positive samples. 
We extracted all positive samples and partial negative samples to get a balanced dataset to evaluate our model.

In the dataset, most textual radiology reports consist of examinations of different body parts. 
Each examination is normally written in the following format. 
It starts with the type of examination, e.g., computed tomography (CT) of the abdomen.  
It is followed by some of the following terms such as clinical history, comparison (to previous examinations), technique, and findings. 
To eliminate textual inputs of unrelated body parts, we extract only the text segment(s) related to hip or pelvis through keywords matching for those documents which follow the format described earlier. 
Afterward, we perform standard text preprocessing steps such as removing punctuation and stopwords as well as converting all alphabetic letters to lower case. 

The dataset is shuffled and then divided into training/development/test sets with 8:1:1 proportion. 
All methods can use development set to tune their hyper-parameters.

\subsection{Target Opinion Word Pairs Extraction}
\subsubsection{Document-level Multi-aspect Sentiment Classification}
\begin{table}[tp]
\centering
\begin{tabular}{l|l|l|l}
\toprule
Dataset & TripAdvisor & BeerAdvocate & HAHip \\ \midrule
\# docs & 28,543 & 27,583 & 5,878 \\
\# target words & 3,737 & 3,088 & 12\\
\# opinion words & 12,406 & 9,166 & 19\\
\# word pairs &208,676               & 249,264   & 11,230  \\
\bottomrule
\end{tabular}
\caption{Statistics of datasets, the vocabulary size of target words and opinion words, and the number of extracted target-opinion word pairs.}
\label{table:as_pair}
\end{table}

In sentiment analysis task, target-opinion word pairs extraction is a well studied problem in sentiment analysis ~\cite{hu2004mining,popescu2007extracting,bloom2007extracting,qiu2011opinion}. 
We designed five rules to extract potential target-opinion word pairs. Our method relies on Stanford Dependency Parser \cite{chen2014fast}. We describe our rules as follows. 

\noindent\textbf{Rule 1}: We extract pairs satisfying the grammatical relation \textit{amod} (adjectival modifier)~\cite{de2008stanford}. For example, in phrase ``very good price,'' we extract ``price'' and ``good'' as a target-opinion pair.

\noindent\textbf{Rule 2}: We extract pairs satisfying the grammatical relation \textit{nsubj} (nominal subject), and the head word is an adjective and the tail word is a noun. For example, in a sentence ``The room is small,'' we can extract ``room'' and ``small'' as a target-opinion pair.

\noindent\textbf{Rule 3}: Some verbs are also opinion words and they are informative. We extract pairs satisfying the grammatical relation \textit{dobj} (direct object) when the head word is one of the following four words: ``like'', ``dislike'', ``love'', and ``hate''. For example, in the sentence ``I like the smell,'' we can extract ``smell'' and ``like'' as a target-opinion pair.

\noindent\textbf{Rule 4}: We extract pairs satisfying the grammatical relation \textit{xcomp} (open clausal complement), and the head word is one of the following word: ``seem'',``look'', ``feel'', ``smell'', and ``taste''. For example, in the sentence ``This beer tastes spicy,'' we can extract ``taste'' and ``spicy'' as a target-opinion pair.

\noindent\textbf{Rule 5}: If the sentence contains some adjectives that can implicitly indicate aspects, we manually assign them to the corresponding aspects. According to \cite{lakkaraju2014aspect}, some adjectives serve both as target words and opinion words. For example, in the sentence ``very tasty, and drinkable,'' the previous rules fail to extract any pair. But we know it contains a target-opinion pair, i.e., (taste, tasty). 
Most of these adjectives have the same root form with the aspects they indicated, e.g., ``clean'' (cleanliness), and ``overpriced'' (price).  This kind of adjective can be extracted first and then we can obtain more similar adjectives using word similarities. For example, given ``tasty,'' we could get ``flavorful'' by retrieving similar words.

Table \ref{table:as_pair} shows the statistics of the rule-based extraction on our two datasets. 
The first four rules can be applied to any dataset while the last one is domain dependent which requires human effort to identify these special adjectives. 
In practice, rule 5 can be removed to save human effort. 
The effect of removing rule 5 is shown in experiments.

After extracting potential target-opinion word pairs, we need to assign them to different aspects as supervision signals. We select some seed words to describe each aspect, and then calculate similarities between the extracted target (or opinion) word and seed words, and assign the pair to the aspect where one of its seed words has the highest similarity. 
The similarity we used is the cosine similarity between two word embeddings trained by word2vec \cite{mikolov2013distributed}. 
For example, suppose seed words $\{$``room'', ``bed''$\}$ and $\{$``business'', ``Internet''$\}$ are used to describe the aspect \textit{room} and \textit{business} respectively, and the candidate pair ``pillow - soft'' will be assigned to the aspect \textit{room} if the similarity between ``pillow'' and ``bed'' is highest among all combinations. 
\subsubsection{Hip Fracture Classification} 
\begin{table}[tp] 
  \centering
    \begin{tabular}{lll|ll}
    \toprule
    \multicolumn{3}{c|}{Opinion Words} & \multicolumn{2}{c}{Target Words} \\
    \midrule
    fracture & fractured & fractures & hip & hips \\
    displaced & dislocated & dislocation & joint & joints \\
    rotated & replacement & destruction & femur & femo \\
    normal & unremarkable & degenerative  & inter-trochanter & trochanter \\
    abnormality & suspicious & impacted & angulation & erosion \\
    injury & pain & olique  & femoral & pelvis \\ 
    intact & & & &\\
    \bottomrule
    \end{tabular}
  \caption{Opinion words and target words in HAHip.}
  \label{tab:extracted_words}%
\end{table}%

We do not use dependency parsers to extract word pairs. 
There is a huge number of medical terminology in radiology reports, the accuracy of dependency parsers drops significantly. 
Radiology reports often comprise abbreviations, acronyms, and grammatical errors \cite{hoang2016text}. 
For example, in a report snippet ``r hip fracture noted," they omit the predicate ``is,'' and use ``r'' to replace ``right.'' 
It is difficult to use dependency parsers to extract satisfactory word pairs. 
Hence, we manually define target words and opinion words lexicons. 
The opinion words and target words used in our method are shown in Table \ref{tab:extracted_words}. 
The statistics of extracted word pairs are shown in Table \ref{table:as_pair}. 
The opinion words mostly are adjectives and target words are nouns. These words are selected manually because of their high frequencies and high relevance. 

We do not use dependency parsers, but we assume every pair of two words within a sentence has a syntactic dependency. 
If within a sentence, there is a word which is in the target words lexicon and a word is in the opinion words lexicon, then the pair of two words should be extracted. 
The idea is shown as follows,
Within a sentence, if there are one opinion word and one target word, they are extracted as a word pair. 
If there is more than one opinion word, we select the opinion word that is nearest to the target words to form the pair. 
If there is more than one target word, we select the target word that is nearest to the opinion words to form the pair. 
In our dataset, most sentences in reports are very short. 
For the sentence which is longer than $20$ words, we manually cut it into several sentences and each of them is no longer than $20$ words.

\subsection{Compared Methods}
\subsubsection{Document-level Multi-aspect Sentiment Classification} 
The goal of DMSC is to predict the sentiment polarity (e.g., positive or negative) of each aspect given a document in which there are several sentences and each sentence describes one or more aspects. 
In the document-level multi-aspect sentiment classification task, we compare our model with the following baselines:

\noindent\textbf{Majority} uses the majority of sentiment polarities in
training sets as predictions.

\noindent\textbf{Lexicon} means using an opinion lexicon to assign sentiment polarity to an aspect \cite{read2009weakly,pablos2015v3}.
We combine two popular opinion lexicons used by \cite{hu2004mining} and \cite{wilson2005recognizing} to get a new one.
If an opinion word from extracted pairs is in the positive (negative) lexicon, it votes for positive (negative).
When the opinion word is with a negation word, its polarity will be flipped.
Then, the polarity of an aspect is determined by using majority voting among all opinion words associated with the aspect.
When the number of positive and negative words is equal, we adopt two different ways to resolve it.
For \textbf{Lexicon-R}, it randomly assigns a polarity.
For \textbf{Lexicon-O}, it uses the overall polarity as the prediction.
Since overall polarities can also be missing, for both Lexicon-R and Lexicon-O, we randomly assign a polarity in uncertain cases and report both mean and std based on five trials of random assignments. 

\noindent\textbf{Assign-O} means directly using the overall polarity of a review in the development set or test set as the prediction for each aspect.

\noindent\textbf{LRR} assumes the overall polarity is a weighted sum of the polarity of each aspect \cite{wang2010latent}.
LRR can be regarded as the only existing weakly supervised baseline where both algorithm and source code are available.

\noindent\textbf{BoW-ALL} is a simple softmax classifier using all annotated training data where the input is a bag-of-words feature vector of a document.  

\noindent\textbf{NDMSC-ALL} is the state-of-the-art neural network based model \cite{yin2017document} (\textbf{NDMSC}) in DMSC task using all annotated training data, which serves an upper bound to our method.

\noindent\textbf{NDMSC-O} is to use overall polarities as a supervision signal to train an NDMSC and apply it to the classification task of each aspect at the test time. 

\noindent\textbf{NDMSC-\{50,100,200,500,1000\}} is the NDMSC algorithm \cite{yin2017document} using partial data. 
In order to see our method is comparable to supervised methods using how many labeled data, we use $\{50,100,200,500,1000\}$ annotated data of each aspect to train NDMSC and compare them to our method.
In addition to annotated data for training, there are extra $20\%$ annotated data for validation.
Since the sampled labeled data may vary for different trials, we perform five trials of random sampling and report both mean and std of the results.

For our method, denoted as \textbf{VUSC}, we assume that a target word and a possible value of sentiment polarity $c$ are independent, and assume $C$ follows a uniform distribution. 
Since the vocabulary size of opinion words is large, we use the negative sampling technique to approximate the partition in Eq. (\ref{dwc}). 
Since there are two sentiment polarities, we do not need to use likelihood ratio estimation to approximate the expectation part.

The document representation we used is obtained from NDMSC \cite{yin2017document}. 
They proposed a novel hierarchical iterative attention model in which documents and pseudo aspect related questions are interleaved at both word and sentence-level to learn an aspect-aware document representation. The pseudo aspect related questions are represented by aspect related keywords.
In order to benefit from their aspect-aware representation scheme, we train an NDMSC to extract the document representation using only overall polarities. 
In the iterative attention module, we use the pseudo aspect related keywords of all aspects released by \cite{yin2017document}. 
One can also use document-to-document autoencoders \cite{li2015hierarchical} to generate the document representation.
In this way, our method can get rid of using overall polarities to generate the document representation. 
Hence, unlike LRR, it is not necessary for our method to use overall polarities. 
Here, to have a fair comparison with LRR, we use the overall polarities to generate document representation. 
For our method, we do not know which state is positive and which one is negative at training time, so the Hungarian algorithm \cite{kuhn1955hungarian} is used to resolve the assignment problem at the test time. 

K-means can be a baseline, but in DMSC task, the inputs of K-means are the same for different aspects, hence, the averaged results of different aspects are very bad. Hence, we do not compare with it in DMSC task. 

\subsubsection{Hip Fracture Classification}
In hip fracture classification task, we compare our model with following baselines:

\noindent\textbf{K-means++} \cite{arthur2007k} is identical to standard K-means clustering algorithm \cite{macqueen1967some} except centroids initialization strategy. 
K-means++ chooses initial centroids one after the other in a special manner such that every new centroid is likely to be far away from the existing centroids. 
The number of centroids is set to two.

\noindent\textbf{Lexicon} refers to an unsupervised approach similar to the Lexicon method in DMSC task. 
The idea is as follows.
Each of the extracted opinion words would vote for \textit{positive} if it belongs to \{fractures, fracture, fractured\}, or it would vote for \textit{negative} if it is in \{no fractures, no fracture, no fractured\}. 
If neither of the above two cases happens, the word would vote for either fracture or no-fracture with equal probability. 
When the numbers of the two types of votes are equal, the sentiment polarity of the document is randomly predicted. 
Finally, the majority voting among all corresponding opinion words would determine the classification result.

\noindent\textbf{LR} Logistic Regression uses a logistic function to model the probability of one category in binary classification. 
The training objective is to find a weight vector such that the overall loss such as binary cross-entropy is minimized, with regularization techniques such as Ridge regularization \cite{doi:10.1080/00401706.1970.10488634}. 
The inputs are BoW (tf-idf) representation.
This method is one of the supervised upper bounds to our approach.

\noindent\textbf{SVM} Support Vector Machine \cite{cortes1995support} algorithm constructs a hyperplane that maximizes the margin between the hyperplane itself and the nearest training data point of any category. 
The inputs are BoW (tf-idf) representation.
We compare our approach with linear support vector machine in the experiment, which serves as a supervised upper-bound to our model.

\noindent\textbf{CNN} Convolutional Neural Network \cite{kim2014convolutional} is a  supervised approach for text classification. 
It implicitly extracts informative local features from input document matrix composed of pre-trained word vectors through operations such as convolution and max-pooling. 
Three different filter sizes $\{3,4,5\}$ are applied in our implementation, and a max pooling layer is applied to each convolutional layer, and each convolutional layer has 100 filters. 
Results of both trainable embeddings setting and fixed embeddings setting are reported.

\noindent\textbf{LSTM} Long Short-Term Memory \cite{gers1999learning} model processes data sequentially through incorporating the information of previous steps. 
In the experiment, we adopt vanilla LSTM implementation \cite{gers1999learning},  and the hidden dimension is set to $50$. 
Similar to CNN, we report the performance of LSTM with trainable embeddings and fixed embeddings. 

For our method, denoted as \textbf{VUSC}, we assume that a target word and a possible value of sentiment polarity $c$ are independent, and assume $C$ follows a uniform distribution. 
Since the vocabulary size of opinion words is small, we do not use the negative sampling technique to approximate the partition in Eq. (\ref{dwc}). 
Since there are two sentiment polarities, we do not use likelihood ratio estimation to approximate the expectation part. 
The document representations of our model are BoW, CNN, and LSTM, which are called VUSC(BoW), VUSC(CNN), and VUSC(LSTM) respectively.


\subsection{Results and Analysis} 

\subsubsection{Document-level Multi-aspect Sentiment Classification}
\begin{table*}[t]
    \centering
        \begin{tabular}{l|cccc|cccc}
            \toprule
            {Dataset} & \multicolumn{4}{c}{TripAdvisor} & \multicolumn{4}{|c}{BeerAdvocate} \\
            
            &  \multicolumn{2}{c}{DEV}  & \multicolumn{2}{c}{TEST}  & \multicolumn{2}{|c}{DEV}  & \multicolumn{2}{c}{TEST} \\
            & Mean & Std & Mean & Std & Mean & Std & Mean & Std \\
            \midrule
            Majority & 0.6286 &-- &0.6242&--&0.6739&--&0.6726&--\\
            Lexicon-R & 0.5914 & 0.0021 &    0.5973 & 0.0018& 0.5895&0.0020&    0.5881 & 0.0025 \\
            Lexicon-O &  0.7153 & 0.0012 & 0.7153 &0.0015 & 0.6510 & 0.0023& 0.6510 & 0.0021 \\
            Assign-O &0.7135 &0.0016 &    0.7043 & 0.0020 & 0.6652 & 0.0028 &    0.6570 & 0.0034 \\
            \hline
            NDMSC-O & 0.7091  &--& 0.7064 & --& 0.6386 &-- &0.6493&--\\
            LRR & 0.6915 &0.0045 & 0.6947 & 0.0024 & 0.5976 &0.0110&0.5941 &0.0113\\
            \textbf{VUSC} &0.7577    & 0.0016 & 0.7561 & 0.0012 & 0.7502 & 0.0058& 0.7538 & 0.0066 \\
            \hline
            NDMSC-50 & 0.7255    & 0.0231 & 0.7270 & 0.0204 & 0.7381 & 0.0143 & 0.7442 & 0.0157 \\
            NDMSC-100 & 0.7482 & 0.0083 & 0.7487 & 0.0069 & 0.7443    & 0.0126 & 0.7493 & 0.0145 \\
            NDMSC-200 & 0.7531    & 0.0040 & 0.7550 & 0.0043 &  0.7555    & 0.0096  & 0.7596 & 0.0092 \\
            NDMSC-500 & 0.7604 & 0.0028  & 0.7616 & 0.0040 & 0.7657 & 0.0066 & 0.7713 & 0.0070 \\
            NDMSC-1000 & 0.7631 & 0.0054 & 0.7638 & 0.0042 & 0.7708    & 0.0066  & 0.7787 & 0.0053 \\
            NDMSC-All & 0.8281 & -- & 0.8334 & -- &0.8576 & -- & 0.8635 & --\\
            BoW-All & 0.8027    & --&    0.8029&    --    &0.8069    &    --&0.8089&    -- \\
            \bottomrule
        \end{tabular} 
    \caption{Averaged accuracies on DMSC of unsupervised, weakly supervised, and supervised methods on TripAdvisor and BeerAdvocate.}
    \label{table:comparison}
\end{table*}
We show all results in Table~\ref{table:comparison}, which consists of three blocks, namely, unsupervised, weakly supervised, and supervised methods. 

For unsupervised methods, our method can outperform the majority on both datasets consistently. 
But other weakly supervised methods cannot outperform the majority on BeerAdvocate dataset, which shows these baselines cannot handle unbalanced data well since BeerAdvocate is more unbalanced than TripAdvisor. 
Our method outperforms Lexicon-R and Lexicon-O, which shows that predicting an opinion word based on a target word may be a better way to use target-opinion pairs, compared with performing a lexicon lookup using opinion words from extract pairs.
Good performance of Lexicon-O and Assign-O demonstrates the usefulness of overall polarities in development/test sets. 
NDMSC-O trained with the overall polarities cannot outperform Assign-O since NDMSC-O can only see overall polarities in training set while Assign-O can see overall polarities for both development and test sets and does not involve learning and generalization. 

For weakly supervised methods, LRR is the only open-source baseline in the literature on weakly supervised DMSC, and our method (VUSC) outperforms LRR by \textbf{6}\% and \textbf{16}\% on TripAdvisor and BeerAdvocate datasets. 
NDMSC-O can also be considered as a weakly supervised method because it only uses overall polarities as a supervision signal, and we still outperform it significantly. 
It is interesting that LRR is worse than NDMSC-O. 
We guess that assuming that the overall polarity is a weighted sum of all aspect polarities may not be a good strategy to train each aspect's polarity or the document representation learned by NDMSC is better than the bag-of-words representation.

For supervised block methods, BoW-All and NDMSC-All are both supervised methods using all annotated data, which can be seen as the upper bound of our algorithm.
NDMSC-All outperforms BoW-All, which shows that the document representation based on neural network is better than the bag-of-words representation. 
Hence, we use the neural networks based document representation as the input of the sentiment polarity classifier. 
Our results are comparable to NDMSC-200 on TripAdvisor and NDMSC-100 on BeerAdvocate. 

\subsubsection{Hip Fracture Classification}

\begin{table*}[t!]
  \centering
    \begin{tabular}{l|cccc}
    \toprule
Methods & \multicolumn{2}{c}{Dev} & \multicolumn{2}{c}{Test} \\
& Mean & Std & Mean & Std \\
\midrule[1pt]
\textbf{VUSC}(BoW) & 0.9086 & 0.0066 & 0.9109 & 0.0056 \\
\textbf{VUSC}(CNN) & 0.9521 & 0.0073 & 0.9383 & 0.0069 \\
\textbf{VUSC}(LSTM) & 0.6627 & 0.0365 & 0.6599 & 0.0247 \\
Lexicon & 0.8457 & 0.0104 & 0.8476 & 0.0089\\
K-means++ & 0.8700 & 0.0005 & 0.8576 & 0.0000\\
\midrule[1pt]
CNN (trainable)& 0.9626 & 0.0013 & 0.9647 & 0.0045 \\
CNN (fixed) & 0.9609 & 0.0007 &  0.9647 & 0.0025 \\
LSTM (trainable) & 0.9366 & 0.0102 & 0.9302 & 0.0042 \\
        
LSTM (fixed) & 0.9322 & 0.0058 & 0.9213 & 0.0025\\
        
LR & 0.9675 & 0.0000 & 0.9678 & 0.0000\\

SVM & 0.9675 & 0.0000 &  0.9695  & 0.0000  \\

    \bottomrule
    \end{tabular}
  \caption{Accuracies of unsupervised and supervised methods on hip fracture classification task. Trainable or fixed means the embeddings are trainable or fixed.}
  \label{tab:classification results}
  
\end{table*}

Regarding unsupervised models, our two models i.e., VUSC(BoW) and VUSC(CNN) outperform other unsupervised methods by a large margin. 
For instance, the accuracy of VUSC(CNN) on the test set is \textbf{9}\% higher than that of Lexicon and \textbf{8}\% higher than that of K-means++. 
For our methods, document representation has a large influence on results. 
The best document representation is CNN. 
CNN representation is slightly better than BoW. 
The possible reason is that BoW is a fixed representation while CNN a is trainable representation. 
CNN has a better capacity for feature extraction. 
The results of VUSC(LSTM) are very low. 
One possible reason is that training LSTM is more difficult compared with CNN, and the supervised signal from target-opinion word pairs is not strong enough to supervise the LSTM. 

In terms of supervised methods, the results of CNN, LR, SVM are comparable. 
The results of LSTM are $3\%$ lower than other supervised methods, which also suggests that LSTM is difficult to train well in this dataset. 
Both our two models, i.e., the accuracies of VUSC(BoW) and VUSC(CNN) are slightly ($5\%$ and $3\%$) lower than CNN, LR, and LSTM, which indicates that our approach is very promising.   
For example, the accuracy of VUSC(CNN) (i.e., $0.9383$) is very close to the best score (i.e., $0.9695$ of SVM) and it is even higher than that of both LSTM models with or without embeddings trainable (i.e., $0.9302$ and $0.9213$). 
\subsection{Ablation Study of Extraction Rules}
To evaluate the effects of extraction rules, we performed an ablation study on DMSC task. 
We run our algorithm VUSC with each rule kept or removed over two datasets. 
If no pairs extracted for one aspect in the training set, the accuracy of this aspect will be 0.5, which is a random guess. 
From the Table~\ref{table:ablation} we can see that the rule R1 is the most effective rule for both datasets. Rules R3/R4/R5 are less effective on their own. 
However, as a whole, they can still improve the overall performance.
When considering removing each rule, we found that our algorithm is quite robust, which indicates missing one of the rules may not hurt the performance much. 
Hence, if human labor is a major concern, rule 5 can be discarded.
We found that sometimes removing one rule may even result in better accuracy (e.g., ``-R3'' for BeerAdvocate dataset).
This means this rule may introduce some noises into the objective function.
However, ``-R3'' can result in worse accuracy for TripAdvisor, which means it is still complementary to the other rules for this dataset. 
\begin{table}[t]
    \centering
        \begin{tabular}{c|cc|cc}
            \toprule
            {Dataset} & \multicolumn{2}{c}{TripAdvisor} & \multicolumn{2}{|c}{BeerAdvocate} \\
            Rule & Dev  & Test & Dev  & Test \\
            \midrule
            R1 & 0.7215 & 0.7174 & 0.7220 & 0.7216 \\
            R2 & 0.7172 & 0.7180 & 0.6864 & 0.6936 \\
            R3 & 0.6263 & 0.6187 & 0.6731 & 0.6725 \\
            R4 & 0.6248 & 0.6279 & 0.6724 & 0.6717 \\
            R5 & 0.5902 & 0.5856 & 0.7095 & 0.7066 \\
            - R1 & 0.7538 & 0.7481 & 0.7458 & 0.7474 \\
            - R2 & 0.7342 & 0.7368 & 0.7504 & 0.7529 \\
            - R3 & 0.7418 & 0.7397 & 0.7565 & 0.7558 \\
            - R4 & 0.7424 & 0.7368 & 0.7518 & 0.7507 \\
            - R5 & 0.7448 & 0.7440 & 0.7550 & 0.7548 \\
            \hline
            All & 0.7577 & 0.7561 & 0.7502 & 0.7538 \\
            \bottomrule
        \end{tabular}
  \caption{Averaged accuracies on DMSC. ``R1\;--\;R5'' means only using a rule while ``-R1\;--$\;$-R5'' means removing a rule from all the rules.}
  \label{table:ablation}
\end{table}
\subsection{Case Study}
We carry out two case studies below to empirically uncover more details of our methods on hip fracture classification task.

\subsubsection{Analysis of Mis-classified Samples}

We present two misclassified samples of VUSC(BoW) and VUSC(CNN), where all sensitive information and irrelevant sentences are removed and replaced by \textit{***}.

Regarding VUSC(BoW), it predicts the report saying ``\textit{*** pelvis and r hip fracture. *** pelvic ring is intact. *** hip joint are unremarkable .}'' to be \emph{negative}, which turns out to be wrong. 
According to the report, pelvis and right hip are fractured, but the pelvic ring and hip joint are not. The model might focus on the latter and thus give a wrong prediction. 
Regarding VUSC(BoW), it predicts the reports saying ``\textit{*** no fracture. *** femur possibly healed old injury to right inferior pubic ramus. ***}'' to be \emph{positive}, which is incorrect.
According to the report, the patient has no fracture but has an old injury. 
In most samples, hip injury comes along with hip fracture. 
This method might capture this pattern. 
As for VUSC(CNN), it predicts the report saying ``\textit{*** there is a displaced left intertrochanteric fracture. no other fracture or dislocation identified. ***}'' as \emph{negative}, which is also very likely to be caused by the misleading sentence ``\textit{no other fracture or dislocation identified}''. 
The model may consider it as ``no fracture''. 
In other report stating that ``\textit{*** right hip pain ***}'', the model predicts it to be \emph{positive}, which is wrong. 
According to the report, the patient feels pain in the right hip. 
In most samples, pain comes along with hip fracture. 
The probability of fracture will increase if there is ``pain'' in the report. 
The model misclassifies the report.  
As can be seen from the above two cases, if the report states many areas of the body, and some areas are fine but some are suggesting fractures, then the model could be confused. 

\subsubsection{Analysis of Word Weights}

\begin{table}[t]
  \centering
    \begin{tabular}{ll|ll}
    \toprule
    \multicolumn{2}{c|}{Positive}  &  \multicolumn{2}{c}{Negative}  \\
    \midrule
    1.  hip & 2.  left & 1.  hip & 2.  no \\
             
    3.  clinical & 4.  right & 3.  fracture & 4.  pelvis  \\
      
    5.  fracture & 6.  femur & 5.  seen  & 6.  right \\
    
    7.  noted & 8.  pain & 7.  left  & 8.  noted  \\
      
    9.  injury & 10.  pelvis & 9.  xr   & 10.  clinical \\

    11. neck & 12. xr  & 11. bony  & 12. diagnosis \\
     
    13. ap & 14. ring  & 13. ap &  14. history \\
     
    15. pelvic & 16. diagnosis & 15. femur & 16. lesion \\
     
    17. history & 18. seen  & 17. bone  & 18. intact \\
    
    19. intact & 20. inter-trochanteric & 19. pelvic & 20. definite \\
    \bottomrule
    \end{tabular}
  \caption{Top 20 largest weighted words for postive weight and negative weight in VUSC(BoW).}
  \label{tab:word_weight}
\end{table}

To have a deeper understanding of how our approach elicits crucial information hidden in unstructured text, we investigate the weight vectors (i.e.,$\mathbf{w}_{c}$ in Eq. (\ref{eq:encoder-softmax})) of the hip fracture classifier in VUSC(BoW) model. 
In Table \ref{tab:word_weight}, we show top $20$
largest weighted words in $\mathbf{w}_{\textit{positive}}$ and $\mathbf{w}_{\textit{negative}}$. 
Large values in the $\mathbf{w}_{\textit{positive}}$ ($\mathbf{w}_{\textit{negative}}$) weight vector mean the associated words having stronger influence to encourage the model to predict \textit{positive} (\textit{negative}) category, as they contribute more in the dot product of logit. 
As for words in the left column, some words including ``fracture'', ``injury'' and ``pain'' also have a strong positive correlation with hip fracture, which also aligns well with intuition. 
For other words which seem to be rather neutral (e.g., ``hip'' and ``left''), it is possible that their high frequency leads to their high value in the weight vector.
Regarding words in the right column, some words such as ``no'' and ``intact'' intuitively make sense as they are powerful indicators of having no hip fracture. 
Nonetheless, it is worthwhile to notice that these words might also induce the model to make false predictions. 
For instance, lots of false negative predictions made by VUSC(BoW) model have both ``no'' and ``fracture''(or ``fractured'') inside the text, which causes the model to categorize them as negative. 
But the keyword ``no'' actually does not syntactically modify ``fracture'', just like the sentence ``no lesion, there is fracture'' in one of the false negative samples. 
Therefore, it is essential to design a more sophisticated mechanism to accurately capture the relationships among words in documents, so that the model can understand the document in a more comprehensive and precise way. 

\subsection{Implementation Details}
We implemented our models using TensorFlow \cite{abadi2016}.
For NDMSC and LRR, we used code released by \cite{yin2017document} and \cite{wang2010latent} respectively, and followed their preprocessing steps and optimal settings.

Parameters are updated by using ADADELTA \cite{zeiler2012adadelta}, an adaptive learning rate method.
To avoid overfitting, we impose weight decay and drop out on both classifiers.
The regularization coefficient and drop out rate are set to $10^{-3}$ and $0.3$ respectively.
In DMSC task, the number of negative samples and $\alpha$ in our model are set to $10$ and $0.1$ respectively. 
For both datasets, $\alpha$ is set to $0.1$. 
In the hip fracture classification task, the number of negative samples is set to $10$.
For VUSC(BoW), $\beta$ and $\gamma$ are set to $0.2$ and $0.8$ respectively. 
For VUSC(CNN), $\beta$ and $\gamma$ are set to $0.9$ and $0.6$ respectively. 
For VUSC(LSTM), $\beta$ and $\gamma$ are set to $0.2$ and $0.4$ respectively. 
The range of $\beta$ and $\gamma$ are $[0.1,0.2,...1.0]$.

For each document (and each aspect), multiple target-opinion pairs are extracted.
The opinion word classifier associated with an aspect will predict five target-opinion pairs at a time.
These five target-opinion pairs are selected with bias.
The probability of a pair being selected is proportional to the frequency of the opinion word to the power of $-0.25$.
In this way, opinion words with low frequency are more likely to be selected compared to uniform sampling. 
In DMSC task, in order to initialize both classifiers better, the word embeddings are retrofitted \cite{faruqui2015} using PPDB \cite{ganitkevitch2013ppdb} semantic lexicons. 
The embeddings size is set $200$. 
In the hip fracture classification task, we train word embeddings using word2vec algorithm \cite{mikolov2013distributed} on the whole dataset.  
The embedding size is set to $50$. 
For our method and K-means++, we do not know which state is positive and which one is negative at training time, so the Hungarian algorithm \cite{kuhn1955hungarian} is used to resolve the assignment problem at the test time. 

%% file: related_work.tex
\section{Related Work}

In this section, we review the related work on unsupervised sentiment analysis, document-level multi-aspect sentiment classification, clinical sentiment analysis, target-opinion word pairs extraction, and variational methods.

\paragraph{Unsupervised Sentiment Analysis.} 
Lexicon-based approaches are typical ways to perform unsupervised sentiment analysis. 
These methods use a sentiment lexicon consists of a list of opinion words along with their sentiment orientation scores to determine the overall sentiment of a given text. 
Some methods \cite{missen2009using,tsytsarau2010scalable} used sentiment orientation scores in existing lexicons directly, and aggregated them within a document to determine polarity. 
Some methods developed their own semantic orientation estimation algorithm. 
For example, \cite{turney2002thumbs} first identified phrases in the review and then estimated the semantic orientation of each extracted phrases. 
The semantic orientation of a given phrase is calculated by comparing its similarity to a positive reference word (``excellent'') with its similarity to a negative reference word (``poor'').
This method determined the sentiment polarity based on the average semantic orientation of the phrases extracted from the review. 
\cite{kamps2004using} used the minimum path distance between a phrase and pivot words (“good” and “bad”) in WordNet to estimate the semantic orientation of extracted phrases. 
Another line of works \cite{li2009non,zhou2014sentiment} proposed a constrained non-negative matrix tri-factorization approach to sentiment analysis, and used a sentiment lexicon as prior knowledge. 
In these models, a term-document matrix is approximated by three factors that specify soft membership of terms and documents in one of $k$ classes. 
All three factors are non-negative matrices. 
The first factor is a matrix representing knowledge in the word space, i.e., each row represents the posterior probability of a word belonging to the $k$ classes. 
The second factor is a matrix providing a condensed view of the term-document matrix.
The third factor is a matrix representing knowledge in document space, i.e., each row represents the posterior probability of a document belonging to the $k$ classes.
\cite{li2009non} applied a regularization to encourage that the first factor is close to prior knowledge. 
\cite{zhou2014sentiment} applied a regularization based on an intuition that if two words (or documents) are sufficiently close to each other, they tend to share the same sentiment polarity.

\paragraph{Document-level Multi-Aspect Sentiment Classification.}
\cite{wang2010latent} proposed an LRR model to solve this problem.
LRR assumes the overall polarity is a weighted sum of all aspect polarities which are represented by word frequency features.
LRR needs to use aspect keywords to perform sentence segmentation to generate the representation of each aspect.
To address the limitation of using aspect keywords, LARAM \cite{wang2011latent} assumes that the text content describing a particular aspect is generated by sampling words from a topic model corresponding to the latent aspect.
Both LRR and LARAM can only access to overall polarities in the training data, but not gold standards of aspect polarities.
\cite{meng2018weakly} proposed a weakly supervised text classification method that can take label surface names, class-related keywords, or a few labeled documents as supervision. 
\cite{ramesh2015weakly} developed a weakly supervised joint model to identify aspects and the corresponding sentiment polarities in online courses. They treat aspect (sentiment) related seed words as weak supervision. 
In the DMSC task which is a fine-grained text classification task, the label surface names or keywords for some aspects would be very similar. 
Given that the inputs are the same and the supervisions are similar, weakly supervised models cannot distinguish them. 
So we do not consider them as our baselines. 
\cite{yin2017document} modeled this problem as a machine comprehension problem under a multi-task learning framework.
It also needs aspect keywords to generate aspect-aware document representations.
Moreover, it can access gold standards of aspect polarities and achieved state-of-the-art performance on this task.
Hence, it can serve as an upper bound.
Some sentence-level aspect based sentiment classification methods \cite{wang2016attention,wang2018aspect} can be directly applied to the DMSC task, because they can solve aspect category sentiment classification task. 
For example, given a sentence ``the restaurant is expensive,'' the aspect category sentiment classification task aims to classify the polarity of the aspect category ``price'' to be \textit{negative}. 
The aspect categories are predefined which are the same as the DMSC task. 
Some of them \cite{tang2016effective,tang2016aspect,chen2017recurrent,ma2017interactive} cannot because they are originally designed for aspect term sentiment classification task. 
For example, given a sentence ``I loved their fajitas,'' the aspect term sentiment classification task aims to classify the polarity of the aspect term ``fajitas'' to be \textit{positive}.  
The aspect terms appearing in the sentence should be provided as inputs.

\paragraph{Clinical Sentiment Analysis.} 
Clinical sentiment analysis aims to determine patients' health status given clinical narratives such as nurse letters, discharge summary, and radiology reports or effectiveness of a treatment or medication given medical social media text \cite{denecke2015sentiment,zunic2020sentiment,abualigah2020sentiment}. 
Here we focus on clinical narratives. 
This task is useful because it could assist physicians to have a complete evaluation of patients' health status or performance diagnosis automatically. 
\cite{bui2014learning} proposed a regular expression discovery (RED) algorithm and implemented two text classifiers based on RED to predict smoking status and pain status. 
\cite{yao2019clinical} proposed a rule-based feature engineering algorithm to identify trigger phrases and then trained knowledge-guided convolutional neural networks to predict patient disease status with respect to obesity and 15 of its comorbidities. 
\cite{eben2019distinguishing} annotated psychiatric electronic health records (EHRs) texts at the sentence level and train lexicon-based, semi-supervised, and supervised machine learning models to predict patients' mental health status. 
\cite{wang2019clinical} developed a rule-based NLP algorithm to automatically generate labels for the training data, and then trained machine learning models including Support Vector Machine (SVM), Random Forrest (RF), Multilayer Perceptron Neural Networks (MLPNN), and Convolutional Neural Networks (CNN) to predict the smoking status and proximal femur (hip) fracture condition. 

\paragraph{Target Opinion Word Pairs Extraction.}
There are two kinds of methods, namely, rule-based methods and learning based methods to solve this task. 
Rule-based methods extract target-opinion word pairs by mining the dependency tree paths between target words and opinion words. 
Learning based methods treat this task as a sequence labeling problem, mapping each word to one of the following categories: target, opinion, and other.
\cite{hu2004mining} is one of the earliest rule-based methods to extract target-opinion pairs. An opinion word is restricted to be an adjective. Target words are extracted first, and then an opinion word is linked to its nearest target word to form a pair.
\cite{popescu2007extracting} and \cite{bloom2007extracting} manually designed dependency tree path templates to extract target-opinion pairs. If the path between a target word candidate and an opinion word candidate belongs to the set of path templates, the pair will be extracted. 
\cite{qiu2011opinion} identified dependency paths that link opinion words and targets via a bootstrapping process. 
This method only needs an initial opinion lexicon to start the bootstrapping process. 
\cite{zhuang2006movie} adopted a supervised learning algorithm to learn valid dependency tree path templates, but it requires target-opinion pairs annotations. 
Learning based methods require lots of target-opinion pairs annotations. 
They trained conditional random fields (CRF) \cite{lafferty2001conditional} based models \cite{jakob2010extracting,yang2012extracting,wang2016recursive} or deep neural networks \cite{liu2015fine,wenya2017coupled,li2017deep} to predict the label (target, opinion or other) of each word. 
\cite{jakob2010extracting} and \cite{li2012cross} extracted target-opinion pairs without using any labeled data in the domain of interest, but it needs lots of labeled data in another related domain.
In this paper, we only use very simple rules to extract target-opinion pairs to validate the effectiveness of our approach. 
If better pairs can be extracted, we can further improve our results.

\paragraph{Variational Methods.}
Variational autoencoders~\cite{KingmaW13,rezende2014stochastic} (VAEs) use a neural network to parameterize a probability distribution. 
VAEs consist of an encoder that parameterizes posterior probabilities and a decoder which parameterizes the reconstruction likelihood given a latent variable. 
VAEs inspire many interesting works \cite{titov2014unsupervised,marcheggiani2016discrete,vsuster2016bilingual,zhang2018variational,chen2018variational,zeng2019variational} which are slightly different from VAEs. Their encoders produce a discrete distribution while the encoder in VAEs yields a continuous latent variable. 
\cite{titov2014unsupervised} aimed to solve the semantic role labeling problem. The encoder is essentially a semantic role labeling model that predicts roles given a rich set of syntactic and lexical features. The decoder reconstructs argument fillers given predicted roles.
\cite{marcheggiani2016discrete} aimed to solve unsupervised open domain relation discovery. The encoder is a feature-rich relation extractor, which predicts a semantic relation between two entities. The decoder reconstructs entities relying on the predicted relation.
\cite{vsuster2016bilingual} tried to learn multi-sense word embeddings. The encoder uses bilingual context to choose a sense for a given word. The decoder predicts context words based on the chosen sense and the given word. 
\cite{zhang2018variational} aimed to solve knowledge graph powered question answering. 
Three neural networks are used to parameterize probabilities of a topic entity given a query and an answer, an answer based on a query and a predicted topic, and the topic given the query. 
\cite{chen2018variational} aimed to infer missing links in a knowledge graph. 
Three neural networks are used to parameterize probabilities of a latent path given two entities and a relation, a relation based on two entities and the chosen latent path, and the relation given the latent path. 
\cite{zeng2019variational} aims to solve document-level multi-aspect sentiment classification task. It uses a neural network to parameterize a discrete distribution which is severed as sentiment classifier and a neural network to parameterize probabilities of opinioned words given target words and possible sentiment polarity.

%% file: conclusion.tex
\section{Conclusion}
In this paper, we propose a variational approach to unsupervised sentiment analysis. 
We extract many target-opinion word pairs by using dependency parsers and simple rules. 
These pairs can be supervision signals to predict sentiment polarity. 
Our objective function is to predict an opinion word given a target word. 
After introducing the sentiment polarity of a document as a latent variable, we can learn a sentiment classifier by optimizing the variational lower bound. 
We also impose sophisticated constraints on extracted words as regularization. 
The experiment results show our method can outperform unsupervised baselines in sentiment analysis task on both domains, and our method obtains comparable results to the supervised method with hundreds of labels per aspect in customer reviews domain, and obtains comparable results to supervised methods in clinical narratives domain. 

%% file: main.bbl
\begin{thebibliography}{}

\bibitem[\protect\BCAY{methods~for classification\ \BBA\ analysis
  of~multivariate observations}{doi}{}]{doi:10.1080/00401706.1970.10488634}


\bibitem[\protect\BCAY{Abadi, Barham, Chen, Chen, Davis, Dean, Devin, Ghemawat,
  Irving, Isard, Kudlur, Levenberg, Monga, Moore, Murray, Steiner, Tucker,
  Vasudevan, Warden, Wicke, Yu,\ \BBA\ Zheng}{Abadi et~al.}{2016}]{abadi2016}
Abadi, M., Barham, P., Chen, J., Chen, Z., Davis, A., Dean, J., Devin, M.,
  Ghemawat, S., Irving, G., Isard, M., Kudlur, M., Levenberg, J., Monga, R.,
  Moore, S., Murray, D.~G., Steiner, B., Tucker, P., Vasudevan, V., Warden, P.,
  Wicke, M., Yu, Y., \BBA\ Zheng, X. \BBOP2016\BBCP.
\newblock \BBOQ Tensorflow: A system for large-scale machine learning\BBCQ\
\newblock In {\Bem Proceedings of OSDI}, \BPGS\ 265--283.

\bibitem[\protect\BCAY{Abualigah, Alfar, Shehab,\ \BBA\ Hussein}{Abualigah
  et~al.}{2020}]{abualigah2020sentiment}
Abualigah, L., Alfar, H.~E., Shehab, M., \BBA\ Hussein, A. M.~A.
  \BBOP2020\BBCP.
\newblock \BBOQ Sentiment analysis in healthcare: A brief review\BBCQ\
\newblock In {\Bem Recent Advances in NLP: The Case of Arabic Language}, \BPGS\
  129--141. Springer.

\bibitem[\protect\BCAY{Arthur\ \BBA\ Vassilvitskii}{Arthur\ \BBA\
  Vassilvitskii}{2007}]{arthur2007k}
Arthur, D.\BBACOMMA\  \BBA\ Vassilvitskii, S. \BBOP2007\BBCP.
\newblock \BBOQ k-means++: The advantages of careful seeding\BBCQ\
\newblock In {\Bem Proceedings of SODA}, \BPGS\ 1027--1035.

\bibitem[\protect\BCAY{Bloom, Garg, Argamon, et~al.}{Bloom
  et~al.}{2007}]{bloom2007extracting}
Bloom, K., Garg, N., Argamon, S., et~al. \BBOP2007\BBCP.
\newblock \BBOQ Extracting appraisal expressions.\BBCQ\
\newblock In {\Bem Proceedings of NAACL-HLT}, \BPGS\ 308--315.

\bibitem[\protect\BCAY{Bui\ \BBA\ Zeng-Treitler}{Bui\ \BBA\
  Zeng-Treitler}{2014}]{bui2014learning}
Bui, D. D.~A.\BBACOMMA\  \BBA\ Zeng-Treitler, Q. \BBOP2014\BBCP.
\newblock \BBOQ Learning regular expressions for clinical text
  classification\BBCQ\
\newblock {\Bem Journal of the American Medical Informatics Association}, {\Bem
  21\/}(5), 850--857.

\bibitem[\protect\BCAY{Chen\ \BBA\ Manning}{Chen\ \BBA\
  Manning}{2014}]{chen2014fast}
Chen, D.\BBACOMMA\  \BBA\ Manning, C. \BBOP2014\BBCP.
\newblock \BBOQ A fast and accurate dependency parser using neural
  networks\BBCQ\
\newblock In {\Bem Proceedings of EMNLP}, \BPGS\ 740--750.

\bibitem[\protect\BCAY{Chen, Sun, Bing,\ \BBA\ Yang}{Chen
  et~al.}{2017}]{chen2017recurrent}
Chen, P., Sun, Z., Bing, L., \BBA\ Yang, W. \BBOP2017\BBCP.
\newblock \BBOQ Recurrent attention network on memory for aspect sentiment
  analysis\BBCQ\
\newblock In {\Bem Proceedings of EMNLP}, \BPGS\ 452--461.

\bibitem[\protect\BCAY{Chen, Xiong, Yan,\ \BBA\ Wang}{Chen
  et~al.}{2018}]{chen2018variational}
Chen, W., Xiong, W., Yan, X., \BBA\ Wang, W. \BBOP2018\BBCP.
\newblock \BBOQ Variational knowledge graph reasoning\BBCQ\
\newblock In {\Bem Proceedings of NAACL-HLT}, \BPGS\ 1823--1832.

\bibitem[\protect\BCAY{Cortes\ \BBA\ Vapnik}{Cortes\ \BBA\
  Vapnik}{1995}]{cortes1995support}
Cortes, C.\BBACOMMA\  \BBA\ Vapnik, V. \BBOP1995\BBCP.
\newblock \BBOQ Support-vector networks\BBCQ\
\newblock {\Bem Machine learning}, {\Bem 20\/}(3), 273--297.

\bibitem[\protect\BCAY{De~Marneffe\ \BBA\ Manning}{De~Marneffe\ \BBA\
  Manning}{2008}]{de2008stanford}
De~Marneffe, M.-C.\BBACOMMA\  \BBA\ Manning, C.~D. \BBOP2008\BBCP.
\newblock \BBOQ Stanford typed dependencies manual\BBCQ\
\newblock \BTR, Technical report, Stanford University.

\bibitem[\protect\BCAY{Denecke\ \BBA\ Deng}{Denecke\ \BBA\
  Deng}{2015}]{denecke2015sentiment}
Denecke, K.\BBACOMMA\  \BBA\ Deng, Y. \BBOP2015\BBCP.
\newblock \BBOQ Sentiment analysis in medical settings: New opportunities and
  challenges\BBCQ\
\newblock {\Bem Artificial intelligence in medicine}, {\Bem 64\/}(1), 17--27.

\bibitem[\protect\BCAY{Devlin, Chang, Lee,\ \BBA\ Toutanova}{Devlin
  et~al.}{2019}]{devlin2018bert}
Devlin, J., Chang, M.-W., Lee, K., \BBA\ Toutanova, K. \BBOP2019\BBCP.
\newblock
\newblock \BBOQ Bert: Pre-training of deep bidirectional transformers for
  language understanding\BBCQ, 4171--4186.

\bibitem[\protect\BCAY{Faruqui, Dodge, Jauhar, Dyer, Hovy,\ \BBA\
  Smith}{Faruqui et~al.}{2015}]{faruqui2015}
Faruqui, M., Dodge, J., Jauhar, S.~K., Dyer, C., Hovy, E., \BBA\ Smith, N.~A.
  \BBOP2015\BBCP.
\newblock \BBOQ Retrofitting word vectors to semantic lexicons\BBCQ\
\newblock In {\Bem Proceedings of NAACL-HLT}, \BPGS\ 1606--1615.

\bibitem[\protect\BCAY{Ganitkevitch, Van~Durme,\ \BBA\
  Callison-Burch}{Ganitkevitch et~al.}{2013}]{ganitkevitch2013ppdb}
Ganitkevitch, J., Van~Durme, B., \BBA\ Callison-Burch, C. \BBOP2013\BBCP.
\newblock \BBOQ Ppdb: The paraphrase database\BBCQ\
\newblock In {\Bem Proceedings of NAACL-HLT}, \BPGS\ 758--764.

\bibitem[\protect\BCAY{Gers, Schmidhuber,\ \BBA\ Cummins}{Gers
  et~al.}{1999}]{gers1999learning}
Gers, F.~A., Schmidhuber, J., \BBA\ Cummins, F. \BBOP1999\BBCP.
\newblock
\newblock \BBOQ Learning to forget: Continual prediction with lstm\BBCQ.

\bibitem[\protect\BCAY{Glynn}{Glynn}{1987}]{glynn1987likelilood}
Glynn, P.~W. \BBOP1987\BBCP.
\newblock \BBOQ Likelilood ratio gradient estimation: an overview\BBCQ\
\newblock In {\Bem Proceedings of CWS}, \BPGS\ 366--375.

\bibitem[\protect\BCAY{Holderness, Cawkwell, Bolton, Pustejovsky,\ \BBA\
  Hall}{Holderness et~al.}{2019}]{eben2019distinguishing}
Holderness, E., Cawkwell, P., Bolton, K., Pustejovsky, J., \BBA\ Hall, M.
  \BBOP2019\BBCP.
\newblock \BBOQ Distinguishing clinical sentiment: The importance of domain
  adaptation in psychiatric patient health records\BBCQ\
\newblock {\Bem CoRR}, {\Bem abs/1904.03225}.

\bibitem[\protect\BCAY{Hu\ \BBA\ Liu}{Hu\ \BBA\ Liu}{2004}]{hu2004mining}
Hu, M.\BBACOMMA\  \BBA\ Liu, B. \BBOP2004\BBCP.
\newblock \BBOQ Mining and summarizing customer reviews\BBCQ\
\newblock In {\Bem Proceedings of SIGKDD}, \BPGS\ 168--177.

\bibitem[\protect\BCAY{Jakob\ \BBA\ Gurevych}{Jakob\ \BBA\
  Gurevych}{2010}]{jakob2010extracting}
Jakob, N.\BBACOMMA\  \BBA\ Gurevych, I. \BBOP2010\BBCP.
\newblock \BBOQ Extracting opinion targets in a single- and cross-domain
  setting with conditional random fields\BBCQ\
\newblock In {\Bem Proceedings of EMNLP}, \BPGS\ 1035--1045.

\bibitem[\protect\BCAY{Kamps, Marx, Mokken, De~Rijke, et~al.}{Kamps
  et~al.}{2004}]{kamps2004using}
Kamps, J., Marx, M., Mokken, R.~J., De~Rijke, M., et~al. \BBOP2004\BBCP.
\newblock \BBOQ Using wordnet to measure semantic orientations of
  adjectives.\BBCQ\
\newblock In {\Bem Proceedings of LREC}, \lowercase{\BVOL}~4, \BPGS\
  1115--1118. Citeseer.

\bibitem[\protect\BCAY{Kim}{Kim}{2014}]{kim2014convolutional}
Kim, Y. \BBOP2014\BBCP.
\newblock
\newblock \BBOQ Convolutional neural networks for sentence classification\BBCQ,
  1746--1751.

\bibitem[\protect\BCAY{Kingma\ \BBA\ Welling}{Kingma\ \BBA\
  Welling}{2014}]{KingmaW13}
Kingma, D.~P.\BBACOMMA\  \BBA\ Welling, M. \BBOP2014\BBCP.
\newblock \BBOQ Auto-encoding variational bayes\BBCQ\
\newblock In {\Bem Proceedings of ICLR}.

\bibitem[\protect\BCAY{Kuhn}{Kuhn}{1955}]{kuhn1955hungarian}
Kuhn, H.~W. \BBOP1955\BBCP.
\newblock \BBOQ The hungarian method for the assignment problem\BBCQ\
\newblock {\Bem Naval Research Logistics}, {\Bem 2\/}(1-2), 83--97.

\bibitem[\protect\BCAY{Lafferty, McCallum,\ \BBA\ Pereira}{Lafferty
  et~al.}{2001}]{lafferty2001conditional}
Lafferty, J.~D., McCallum, A., \BBA\ Pereira, F. C.~N. \BBOP2001\BBCP.
\newblock \BBOQ Conditional random fields: Probabilistic models for segmenting
  and labeling sequence data\BBCQ\
\newblock In {\Bem Proceedings of ICML}, \BPGS\ 282--289.

\bibitem[\protect\BCAY{Lakkaraju, Socher,\ \BBA\ Manning}{Lakkaraju
  et~al.}{2014}]{lakkaraju2014aspect}
Lakkaraju, H., Socher, R., \BBA\ Manning, C. \BBOP2014\BBCP.
\newblock \BBOQ Aspect specific sentiment analysis using hierarchical deep
  learning\BBCQ\
\newblock In {\Bem Proceedings of NeurIPS workshop on Deep Learning and
  Representation Learning}.

\bibitem[\protect\BCAY{Lei, Barzilay,\ \BBA\ Jaakkola}{Lei
  et~al.}{2016}]{lei2016rationalizing}
Lei, T., Barzilay, R., \BBA\ Jaakkola, T. \BBOP2016\BBCP.
\newblock \BBOQ Rationalizing neural predictions\BBCQ\
\newblock In {\Bem Proceedings of EMNLP}, \BPGS\ 107--117.

\bibitem[\protect\BCAY{Li, Pan, Jin, Yang,\ \BBA\ Zhu}{Li
  et~al.}{2012}]{li2012cross}
Li, F., Pan, S.~J., Jin, O., Yang, Q., \BBA\ Zhu, X. \BBOP2012\BBCP.
\newblock \BBOQ Cross-domain co-extraction of sentiment and topic
  lexicons\BBCQ\
\newblock In {\Bem Proceedings of ACL}, \BPGS\ 410--419.

\bibitem[\protect\BCAY{Li, Luong,\ \BBA\ Jurafsky}{Li
  et~al.}{2015}]{li2015hierarchical}
Li, J., Luong, M.-T., \BBA\ Jurafsky, D. \BBOP2015\BBCP.
\newblock \BBOQ A hierarchical neural autoencoder for paragraphs and
  documents\BBCQ\
\newblock In {\Bem Proceedings of ACL}, \BPGS\ 1106--1115.

\bibitem[\protect\BCAY{Li, Zhang,\ \BBA\ Sindhwani}{Li
  et~al.}{2009}]{li2009non}
Li, T., Zhang, Y., \BBA\ Sindhwani, V. \BBOP2009\BBCP.
\newblock \BBOQ A non-negative matrix tri-factorization approach to sentiment
  classification with lexical prior knowledge\BBCQ\
\newblock In {\Bem Proceedings of ACL}, \BPGS\ 244--252.

\bibitem[\protect\BCAY{Li\ \BBA\ Lam}{Li\ \BBA\ Lam}{2017}]{li2017deep}
Li, X.\BBACOMMA\  \BBA\ Lam, W. \BBOP2017\BBCP.
\newblock \BBOQ Deep multi-task learning for aspect term extraction with memory
  interaction\BBCQ\
\newblock In {\Bem Proceedings of EMNLP}, \BPGS\ 2886--2892.

\bibitem[\protect\BCAY{Liu}{Liu}{2012}]{liu2012sentiment}
Liu, B. \BBOP2012\BBCP.
\newblock \BBOQ Sentiment analysis and opinion mining\BBCQ\
\newblock {\Bem Synthesis lectures on human language technologies}, {\Bem
  5\/}(1), 1--167.

\bibitem[\protect\BCAY{Liu, Joty,\ \BBA\ Meng}{Liu et~al.}{2015}]{liu2015fine}
Liu, P., Joty, S., \BBA\ Meng, H. \BBOP2015\BBCP.
\newblock \BBOQ Fine-grained opinion mining with recurrent neural networks and
  word embeddings\BBCQ\
\newblock In {\Bem Proceedings of EMNLP}, \BPGS\ 1433--1443.

\bibitem[\protect\BCAY{Ma, Li, Zhang,\ \BBA\ Wang}{Ma
  et~al.}{2017}]{ma2017interactive}
Ma, D., Li, S., Zhang, X., \BBA\ Wang, H. \BBOP2017\BBCP.
\newblock \BBOQ Interactive attention networks for aspect-level sentiment
  classification\BBCQ\
\newblock In {\Bem Proceedings of IJCAI}, \BPGS\ 4068--4074.

\bibitem[\protect\BCAY{MacQueen et~al.}{MacQueen
  et~al.}{1967}]{macqueen1967some}
MacQueen, J.\BBACOMMA\  et~al. \BBOP1967\BBCP.
\newblock \BBOQ Some methods for classification and analysis of multivariate
  observations\BBCQ\
\newblock In {\Bem Proceedings of the fifth Berkeley symposium on mathematical
  statistics and probability}, \lowercase{\BVOL}~1, \BPGS\ 281--297. Oakland,
  CA, USA.

\bibitem[\protect\BCAY{Marcheggiani\ \BBA\ Titov}{Marcheggiani\ \BBA\
  Titov}{2016}]{marcheggiani2016discrete}
Marcheggiani, D.\BBACOMMA\  \BBA\ Titov, I. \BBOP2016\BBCP.
\newblock \BBOQ Discrete-state variational autoencoders for joint discovery and
  factorization of relations\BBCQ\
\newblock {\Bem Transactions of the Association for Computational Linguistics},
  {\Bem 4}, 231--244.

\bibitem[\protect\BCAY{McAuley, Leskovec,\ \BBA\ Jurafsky}{McAuley
  et~al.}{2012}]{mcauley2012learning}
McAuley, J., Leskovec, J., \BBA\ Jurafsky, D. \BBOP2012\BBCP.
\newblock \BBOQ Learning attitudes and attributes from multi-aspect
  reviews\BBCQ\
\newblock In {\Bem Proceedings of ICDM}, \BPGS\ 1020--1025.

\bibitem[\protect\BCAY{Meng, Shen, Zhang,\ \BBA\ Han}{Meng
  et~al.}{2018}]{meng2018weakly}
Meng, Y., Shen, J., Zhang, C., \BBA\ Han, J. \BBOP2018\BBCP.
\newblock \BBOQ Weakly-supervised neural text classification\BBCQ\
\newblock In {\Bem Proceedings of CIKM}, \BPGS\ 983--992.

\bibitem[\protect\BCAY{Mikolov, Sutskever, Chen, Corrado,\ \BBA\ Dean}{Mikolov
  et~al.}{2013}]{mikolov2013distributed}
Mikolov, T., Sutskever, I., Chen, K., Corrado, G.~S., \BBA\ Dean, J.
  \BBOP2013\BBCP.
\newblock \BBOQ Distributed representations of words and phrases and their
  compositionality\BBCQ\
\newblock In {\Bem Proceedings of NeurIPS}, \BPGS\ 3111--3119.

\bibitem[\protect\BCAY{Missen\ \BBA\ Boughanem}{Missen\ \BBA\
  Boughanem}{2009}]{missen2009using}
Missen, M. M.~S.\BBACOMMA\  \BBA\ Boughanem, M. \BBOP2009\BBCP.
\newblock \BBOQ Using wordnet’s semantic relations for opinion detection in
  blogs\BBCQ\
\newblock In {\Bem Proceedings of ECIR}, \BPGS\ 729--733. Springer.

\bibitem[\protect\BCAY{Ng, Dasgupta,\ \BBA\ Arifin}{Ng
  et~al.}{2006}]{ng2006examining}
Ng, V., Dasgupta, S., \BBA\ Arifin, S.~N. \BBOP2006\BBCP.
\newblock \BBOQ Examining the role of linguistic knowledge sources in the
  automatic identification and classification of reviews\BBCQ\
\newblock In {\Bem Proceedings of ACL}, \BPGS\ 611--618.

\bibitem[\protect\BCAY{Nguyen\ \BBA\ Patrick}{Nguyen\ \BBA\
  Patrick}{2016}]{hoang2016text}
Nguyen, H.\BBACOMMA\  \BBA\ Patrick, J. \BBOP2016\BBCP.
\newblock \BBOQ Text mining in clinical domain: Dealing with noise\BBCQ\
\newblock In Krishnapuram, B., Shah, M., Smola, A.~J., Aggarwal, C.~C., Shen,
  D., \BBA\ Rastogi, R.\BEDS, {\Bem Proceedings of SIGKDD}, \BPGS\ 549--558.

\bibitem[\protect\BCAY{Pablos, Cuadros,\ \BBA\ Rigau}{Pablos
  et~al.}{2015}]{pablos2015v3}
Pablos, A.~G., Cuadros, M., \BBA\ Rigau, G. \BBOP2015\BBCP.
\newblock \BBOQ V3: Unsupervised aspect based sentiment analysis for
  semeval2015 task 12\BBCQ\
\newblock In {\Bem Proceedings of SemEval}, \BPGS\ 714--718.

\bibitem[\protect\BCAY{Popescu\ \BBA\ Etzioni}{Popescu\ \BBA\
  Etzioni}{2005}]{popescu2007extracting}
Popescu, A.-M.\BBACOMMA\  \BBA\ Etzioni, O. \BBOP2005\BBCP.
\newblock \BBOQ Extracting product features and opinions from reviews\BBCQ\
\newblock In {\Bem Proceedings of EMNLP-HLT}, \BPGS\ 339--346.

\bibitem[\protect\BCAY{Qiu, Liu, Bu,\ \BBA\ Chen}{Qiu
  et~al.}{2011}]{qiu2011opinion}
Qiu, G., Liu, B., Bu, J., \BBA\ Chen, C. \BBOP2011\BBCP.
\newblock \BBOQ Opinion word expansion and target extraction through double
  propagation\BBCQ\
\newblock {\Bem Computational linguistics}, {\Bem 37}, 9--27.

\bibitem[\protect\BCAY{Ramesh, Kumar, Foulds,\ \BBA\ Getoor}{Ramesh
  et~al.}{2015}]{ramesh2015weakly}
Ramesh, A., Kumar, S.~H., Foulds, J., \BBA\ Getoor, L. \BBOP2015\BBCP.
\newblock \BBOQ Weakly supervised models of aspect-sentiment for online course
  discussion forums\BBCQ\
\newblock In {\Bem Proceedings of ACL}, \BPGS\ 74--83.

\bibitem[\protect\BCAY{Read\ \BBA\ Carroll}{Read\ \BBA\
  Carroll}{2009}]{read2009weakly}
Read, J.\BBACOMMA\  \BBA\ Carroll, J. \BBOP2009\BBCP.
\newblock \BBOQ Weakly supervised techniques for domain-independent sentiment
  classification\BBCQ\
\newblock In {\Bem Proceedings of CIKM workshop on Topic-sentiment Analysis for
  Mass Opinion}, \BPGS\ 45--52.

\bibitem[\protect\BCAY{Rezende, Mohamed,\ \BBA\ Wierstra}{Rezende
  et~al.}{2014}]{rezende2014stochastic}
Rezende, D.~J., Mohamed, S., \BBA\ Wierstra, D. \BBOP2014\BBCP.
\newblock \BBOQ Stochastic backpropagation and approximate inference in deep
  generative models\BBCQ\
\newblock In {\Bem Proceedings of ICML}, \BPGS\ 1278--1286.

\bibitem[\protect\BCAY{{\v{S}}uster, Titov,\ \BBA\ van Noord}{{\v{S}}uster
  et~al.}{2016}]{vsuster2016bilingual}
{\v{S}}uster, S., Titov, I., \BBA\ van Noord, G. \BBOP2016\BBCP.
\newblock \BBOQ Bilingual learning of multi-sense embeddings with discrete
  autoencoders\BBCQ\
\newblock In {\Bem Proceedings of NAACL-HLT}, \BPGS\ 1346--1356.

\bibitem[\protect\BCAY{Taboada, Brooke, Tofiloski, Voll,\ \BBA\ Stede}{Taboada
  et~al.}{2011}]{maite2011lexicon}
Taboada, M., Brooke, J., Tofiloski, M., Voll, K.~D., \BBA\ Stede, M.
  \BBOP2011\BBCP.
\newblock \BBOQ Lexicon-based methods for sentiment analysis\BBCQ\
\newblock {\Bem Computational Linguistics}, {\Bem 37\/}(2), 267--307.

\bibitem[\protect\BCAY{Tang, Qin, Feng,\ \BBA\ Liu}{Tang
  et~al.}{2016a}]{tang2016effective}
Tang, D., Qin, B., Feng, X., \BBA\ Liu, T. \BBOP2016a\BBCP.
\newblock \BBOQ Effective lstms for target-dependent sentiment
  classification\BBCQ\
\newblock In {\Bem Proceedings of COLING}, \BPGS\ 3298--3307.

\bibitem[\protect\BCAY{Tang, Qin,\ \BBA\ Liu}{Tang
  et~al.}{2016b}]{tang2016aspect}
Tang, D., Qin, B., \BBA\ Liu, T. \BBOP2016b\BBCP.
\newblock \BBOQ Aspect level sentiment classification with deep memory
  network\BBCQ\
\newblock In {\Bem Proceedings of EMNLP}, \BPGS\ 214--224.

\bibitem[\protect\BCAY{Titov\ \BBA\ Khoddam}{Titov\ \BBA\
  Khoddam}{2015}]{titov2014unsupervised}
Titov, I.\BBACOMMA\  \BBA\ Khoddam, E. \BBOP2015\BBCP.
\newblock \BBOQ Unsupervised induction of semantic roles within a
  reconstruction-error minimization framework\BBCQ\
\newblock In {\Bem Proceedings of NAACL-HLT}, \BPGS\ 1--10.

\bibitem[\protect\BCAY{Tsytsarau, Palpanas,\ \BBA\ Denecke}{Tsytsarau
  et~al.}{2010}]{tsytsarau2010scalable}
Tsytsarau, M., Palpanas, T., \BBA\ Denecke, K. \BBOP2010\BBCP.
\newblock \BBOQ Scalable discovery of contradictions on the web\BBCQ\
\newblock In {\Bem Proceedings of WWW}, \BPGS\ 1195--1196.

\bibitem[\protect\BCAY{Turney}{Turney}{2002}]{turney2002thumbs}
Turney, P.~D. \BBOP2002\BBCP.
\newblock \BBOQ Thumbs up or thumbs down?: semantic orientation applied to
  unsupervised classification of reviews\BBCQ\
\newblock In {\Bem Proceedings of ACL}, \BPGS\ 417--424.

\bibitem[\protect\BCAY{Wang, Lu,\ \BBA\ Zhai}{Wang
  et~al.}{2010}]{wang2010latent}
Wang, H., Lu, Y., \BBA\ Zhai, C. \BBOP2010\BBCP.
\newblock \BBOQ Latent aspect rating analysis on review text data: a rating
  regression approach\BBCQ\
\newblock In {\Bem Proceedings of SIGKDD}, \BPGS\ 783--792.

\bibitem[\protect\BCAY{Wang, Lu,\ \BBA\ Zhai}{Wang
  et~al.}{2011}]{wang2011latent}
Wang, H., Lu, Y., \BBA\ Zhai, C. \BBOP2011\BBCP.
\newblock \BBOQ Latent aspect rating analysis without aspect keyword
  supervision\BBCQ\
\newblock In {\Bem Proceedings of SIGKDD}, \BPGS\ 618--626.

\bibitem[\protect\BCAY{Wang, Li, Li, Kang, Zhang, Si,\ \BBA\ Zhou}{Wang
  et~al.}{2018}]{wang2018aspect}
Wang, J., Li, J., Li, S., Kang, Y., Zhang, M., Si, L., \BBA\ Zhou, G.
  \BBOP2018\BBCP.
\newblock \BBOQ Aspect sentiment classification with both word-level and
  clause-level attention networks.\BBCQ\
\newblock In {\Bem Proceedings of IJCAI}, \BPGS\ 4439--4445.

\bibitem[\protect\BCAY{Wang, Pan, Dahlmeier,\ \BBA\ Xiao}{Wang
  et~al.}{2016}]{wang2016recursive}
Wang, W., Pan, S.~J., Dahlmeier, D., \BBA\ Xiao, X. \BBOP2016\BBCP.
\newblock \BBOQ Recursive neural conditional random fields for aspect-based
  sentiment analysis\BBCQ\
\newblock In {\Bem Proceedings of EMNLP}, \BPGS\ 616--626.

\bibitem[\protect\BCAY{Wang, Pan, Dahlmeier,\ \BBA\ Xiao}{Wang
  et~al.}{2017}]{wenya2017coupled}
Wang, W., Pan, S.~J., Dahlmeier, D., \BBA\ Xiao, X. \BBOP2017\BBCP.
\newblock \BBOQ Coupled multi-layer attentions for co-extraction of aspect and
  opinion terms\BBCQ\
\newblock In {\Bem Proceedings of AAAI}, \BPGS\ 3316--3322.

\bibitem[\protect\BCAY{Wang, Sohn, Liu, Shen, Wang, Atkinson, Amin,\ \BBA\
  Liu}{Wang et~al.}{2019}]{wang2019clinical}
Wang, Y., Sohn, S., Liu, S., Shen, F., Wang, L., Atkinson, E.~J., Amin, S.,
  \BBA\ Liu, H. \BBOP2019\BBCP.
\newblock \BBOQ A clinical text classification paradigm using weak supervision
  and deep representation\BBCQ\
\newblock {\Bem BMC Medical Informatics and Decision Making}, {\Bem 19\/}(1),
  1--13.

\bibitem[\protect\BCAY{Wang, Huang, Zhao, et~al.}{Wang
  et~al.}{2016}]{wang2016attention}
Wang, Y., Huang, M., Zhao, L., et~al. \BBOP2016\BBCP.
\newblock \BBOQ Attention-based lstm for aspect-level sentiment
  classification\BBCQ\
\newblock In {\Bem Proceedings of EMNLP}, \BPGS\ 606--615.

\bibitem[\protect\BCAY{Wilson, Wiebe,\ \BBA\ Hoffmann}{Wilson
  et~al.}{2005}]{wilson2005recognizing}
Wilson, T., Wiebe, J., \BBA\ Hoffmann, P. \BBOP2005\BBCP.
\newblock \BBOQ Recognizing contextual polarity in phrase-level sentiment
  analysis\BBCQ\
\newblock In {\Bem Proceedings of EMNLP-HLT}, \BPGS\ 347--354.

\bibitem[\protect\BCAY{Yang\ \BBA\ Cardie}{Yang\ \BBA\
  Cardie}{2012}]{yang2012extracting}
Yang, B.\BBACOMMA\  \BBA\ Cardie, C. \BBOP2012\BBCP.
\newblock \BBOQ Extracting opinion expressions with semi-markov conditional
  random fields\BBCQ\
\newblock In {\Bem Proceedings of EMNLP-CoNLL}, \BPGS\ 1335--1345.

\bibitem[\protect\BCAY{Yao, Mao,\ \BBA\ Luo}{Yao
  et~al.}{2019}]{yao2019clinical}
Yao, L., Mao, C., \BBA\ Luo, Y. \BBOP2019\BBCP.
\newblock \BBOQ Clinical text classification with rule-based features and
  knowledge-guided convolutional neural networks\BBCQ\
\newblock {\Bem BMC Medical Informatics and Decision Making}, {\Bem 19-S\/}(3),
  31--39.

\bibitem[\protect\BCAY{Yin, Song,\ \BBA\ Zhang}{Yin
  et~al.}{2017}]{yin2017document}
Yin, Y., Song, Y., \BBA\ Zhang, M. \BBOP2017\BBCP.
\newblock \BBOQ Document-level multi-aspect sentiment classification as machine
  comprehension\BBCQ\
\newblock In {\Bem Proceedings of EMNLP}, \BPGS\ 2034--2044.

\bibitem[\protect\BCAY{Zeiler}{Zeiler}{2012}]{zeiler2012adadelta}
Zeiler, M.~D. \BBOP2012\BBCP.
\newblock \BBOQ Adadelta: an adaptive learning rate method\BBCQ\
\newblock {\Bem arXiv preprint arXiv:1212.5701}.

\bibitem[\protect\BCAY{Zeng, Zhou, Liu,\ \BBA\ Song}{Zeng
  et~al.}{2019}]{zeng2019variational}
Zeng, Z., Zhou, W., Liu, X., \BBA\ Song, Y. \BBOP2019\BBCP.
\newblock \BBOQ A variational approach to weakly supervised document-level
  multi-aspect sentiment classification\BBCQ\
\newblock In {\Bem Proceedings of NAACL-HLT}, \BPGS\ 386--396.

\bibitem[\protect\BCAY{Zhang, Dai, Kozareva, Smola,\ \BBA\ Song}{Zhang
  et~al.}{2018}]{zhang2018variational}
Zhang, Y., Dai, H., Kozareva, Z., Smola, A.~J., \BBA\ Song, L. \BBOP2018\BBCP.
\newblock \BBOQ Variational reasoning for question answering with knowledge
  graph\BBCQ\
\newblock In {\Bem Proceedings of AAAI}, \BPGS\ 6069--6076.

\bibitem[\protect\BCAY{Zhou, Zhao,\ \BBA\ Zeng}{Zhou
  et~al.}{2014}]{zhou2014sentiment}
Zhou, G., Zhao, J., \BBA\ Zeng, D. \BBOP2014\BBCP.
\newblock \BBOQ Sentiment classification with graph co-regularization\BBCQ\
\newblock In {\Bem Proceedings of COLING}, \BPGS\ 1331--1340.

\bibitem[\protect\BCAY{Zhuang, Jing,\ \BBA\ Zhu}{Zhuang
  et~al.}{2006}]{zhuang2006movie}
Zhuang, L., Jing, F., \BBA\ Zhu, X.-Y. \BBOP2006\BBCP.
\newblock \BBOQ Movie review mining and summarization\BBCQ\
\newblock In {\Bem Proceedings of CIKM}, \BPGS\ 43--50.

\bibitem[\protect\BCAY{Zunic, Corcoran,\ \BBA\ Spasic}{Zunic
  et~al.}{2020}]{zunic2020sentiment}
Zunic, A., Corcoran, P., \BBA\ Spasic, I. \BBOP2020\BBCP.
\newblock \BBOQ Sentiment analysis in health and well-being: Systematic
  review\BBCQ\
\newblock {\Bem JMIR Medical Informatics}, {\Bem 8\/}(1), e16023.

\end{thebibliography}
